\ifdefined\pdfsuppressptexinfo
\fi

\documentclass[10pt]{article}
\usepackage[preprint]{tmlr}
\usepackage{amsmath}
\usepackage{amssymb}
\usepackage{algorithm}
\usepackage{algorithmic}
\usepackage{array}
\usepackage{graphicx}
\usepackage{hyperref}
\usepackage{url}

\hypersetup{
  unicode=true,
  pdftitle={MORPH: Generative Retrieval via Diffusion Transformer with Metric-Ordered Sequence Training and Hybrid-Policy Preference Optimization},
  pdfauthor={Chenghao Liu, Yu Zhang, Zhongtao Jiang, Kun X\'{u}, Zhenwei An, Renzhi Wang, Zhao Wang, Jiachen Zhang, Yuxiao Zhang, Kun X\v{u}, Songfang Huang},
  pdfkeywords={},
  pdfcreator={},
  pdfproducer={}
}

\title{MORPH: Generative Retrieval via Diffusion Transformer with Metric-Ordered Sequence Training and Hybrid-Policy Preference Optimization}

\author{%
  \parbox{\textwidth}{\centering
    {\name Chenghao~Liu$^{*}$, Yu~Zhang$^{*}$, Zhongtao~Jiang,
      Kun~X\'{u}, Zhenwei~An, Renzhi~Wang\\
      Zhao~Wang, Jiachen~Zhang, Yuxiao~Zhang,
      Kun~X\v{u}$^{\dagger}$, Songfang~Huang$^{\dagger}$}\\[6pt]
    {\addr\upshape\bfseries Peking University}}}

\newcommand{\morphfullname}{Generative \textbf{R}etrieval via Diffusion Transformer with
  \textbf{M}etric-\textbf{O}rdered Sequence Training and \textbf{H}ybrid-Policy
  \textbf{P}reference Optimization}

\newcommand{\method}{MORPH}
\newcommand{\mos}{MOS}
\newcommand{\hppo}{HPPO}
\newcommand{\ret}{\mathcal{R}}
\newcommand{\ind}{\mathbb{I}}

\newcommand{\metricIssue}{\texttt{Attr@K}}
\newcommand{\metricSame}{\texttt{Same@K}}
\newcommand{\metricJoint}{\texttt{Joint@K}}
\newcommand{\metricCond}{\texttt{Cond@K}}
\newcommand{\shortJoint}{\texttt{Joint@K}}

\newcommand{\shortSame}{\texttt{Same@K}}

\begin{document}

\maketitle
\let\AND\relax

\makeatletter
\long\def\@makefntext#1{\parindent\z@\noindent\@makefnmark#1}
\makeatother

\begingroup
\renewcommand{\thefootnote}{}
\footnotetext{\raggedright $^{*}$Equal contribution.\quad $^{\dagger}$Corresponding authors.\par
  Emails:
  \mbox{\nolinkurl{chliu@stu.pku.edu.cn}}, \mbox{\nolinkurl{zhangyuthu14@gmail.com}},
  \mbox{\nolinkurl{ji4ngxiaoming@gmail.com}}, \mbox{\nolinkurl{xukunxkxk@gmail.com}},
  \mbox{\nolinkurl{anzhenwei@pku.edu.cn}}, \mbox{\nolinkurl{rzhwang.cn@gmail.com}},
  \mbox{\nolinkurl{zhao.wang.scolar@gmail.com}}, \mbox{\nolinkurl{z89498323286@gmail.com}},
  \mbox{\nolinkurl{yuxiao_zh@foxmail.com}}, \mbox{\nolinkurl{syxu828@gmail.com}},
  \mbox{\nolinkurl{hsf@pku.edu.cn}}.\par}
\endgroup

\begin{abstract}
Embedding-based retrieval typically returns highest-scoring items, but many production scenarios require items that satisfy a target attribute while preserving a fine-grained pattern expressed by seed examples.\footnote{We used GPT to polish the manuscript.} We formalize this as \emph{pattern-preserving attribute retrieval}. Standard approaches fail: averaging seeds preserves the pattern but misses the attribute; global attribute retrieval drifts to unrelated patterns. We approach the task with continuous generative retrieval, where a model reads item-embedding sequences and generates query embeddings for nearest-neighbor search. We propose \textbf{\method{}}: \morphfullname{}, a staged framework with large-scale raw-sequence pretraining, Metric-Ordered Sequence (\mos{}) training, and final \hppo{} alignment. \mos{} construction turns sparse online metric labels into in-pattern trajectories; \mos{} CPT/SFT then trains the generator through shared multi-domain continuation pretraining and domain-specific tail-centroid supervised fine-tuning. \hppo{} uses a hybrid pool of static and policy-generated candidate embeddings, labels them with true online intersection metrics, applies iterated preference optimization, and employs a Pareto pair filter to exclude winners that lower pattern purity. Across four large-scale attribute domains under strict item- and pattern-holdout protocols, \mos{} training improves the primary intersection metric over a strong pretrained generative retriever in every domain--split cell, and the complete Pareto-filtered \hppo{} procedure improves it further, with paired-bootstrap-significant gains on seven of the eight cells---the exception being the D4 pattern-holdout split. Ablations confirm that the Pareto pair filter improves the attribute--pattern tradeoff on D1--D3, and that hybrid static/policy candidates are complementary.
\end{abstract}

\section{Introduction}

\begin{figure}[!t]
\centering
\includegraphics[width=\textwidth]{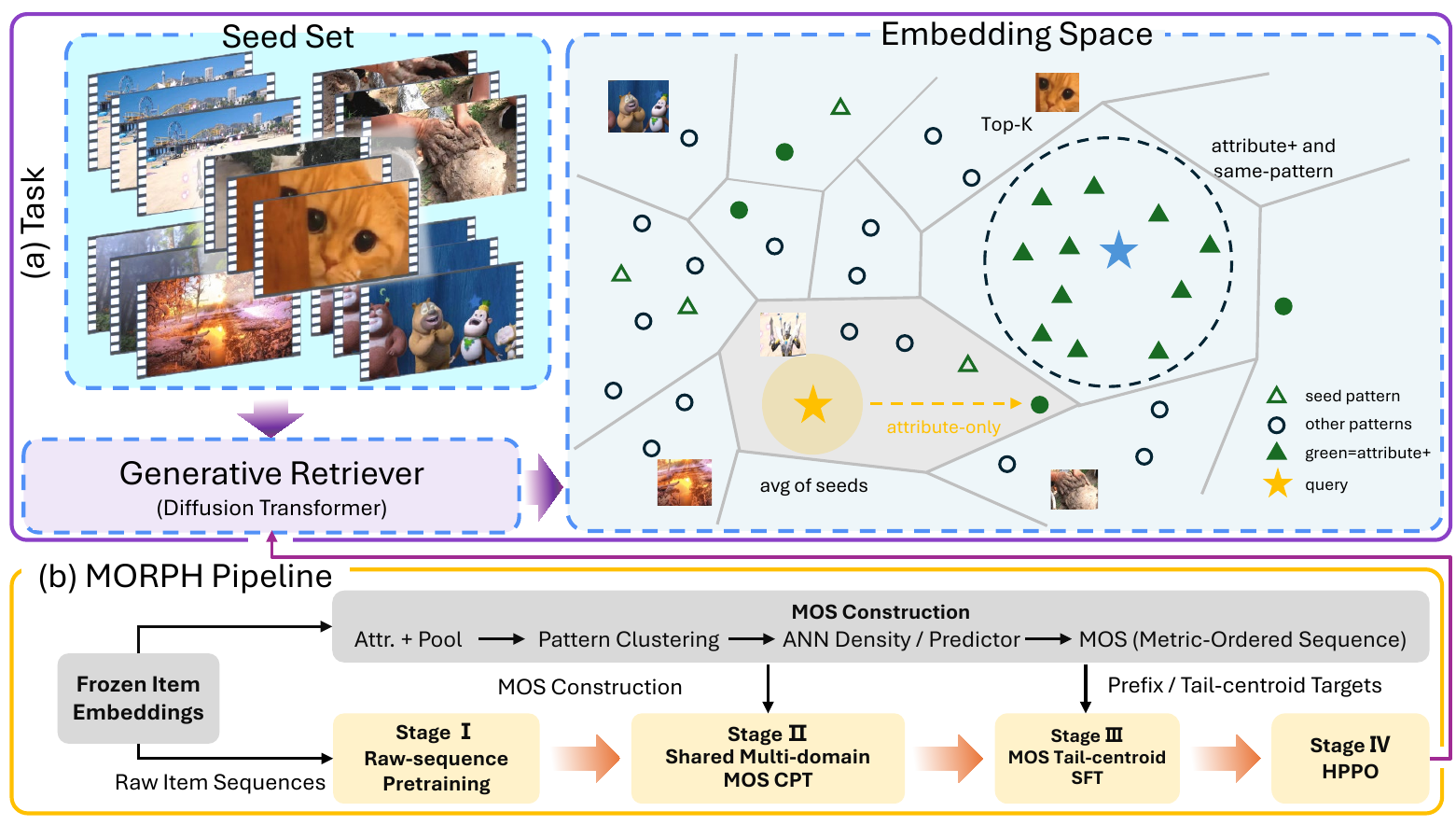}
\caption{
\textbf{Pattern-preserving attribute retrieval and the \method{} pipeline.}
The upper panel illustrates the task: given a seed set from one fine-grained pattern, the generative retriever synthesizes a continuous query embedding whose top-$K$ neighborhood has a high density of items that are both attribute-positive and from the seed pattern. Averaging the seed embeddings preserves the pattern but remains in a low-attribute-density region, whereas attribute-only retrieval may drift toward unrelated patterns. The lower panel separates offline data construction from four weight-updating stages. Frozen item embeddings provide raw item sequences for Stage~I raw-sequence pretraining and an attribute-positive item pool for offline \mos{} construction. Pattern clustering and ANN-density prediction are used to construct metric-ordered sequences, which supervise Stage~II \mos{} CPT and provide prefix/tail-centroid targets for Stage~III \mos{} SFT. Stage~IV HPPO further aligns the resulting retriever with the true online pattern-preserving retrieval objective. Item thumbnails are sampled from a public dataset~\citep{ni2023content} for illustration only.
}
\label{fig:task}
\end{figure}

Large retrieval systems increasingly serve attribute-seeking objectives. A platform may need to find items that satisfy a quality, safety, policy, originality, or integrity criterion. In many such tasks, the desired result is not an arbitrary high-scoring item. Retrieved items should also match the fine-grained pattern represented by a seed set: the same topic, visual style, failure mode, or retrieval intent. A typical operational case illustrates the tension: reviewers discover a small batch of items that express one new variant of a policy violation---a specific format, template, or tactic---and need to retrieve many more items of exactly that variant for measurement and treatment. A global attribute classifier ranks items by the target attribute without conditioning on the seed pattern; for a fixed attribute, it therefore produces the same ranking for every seed set and cannot restrict retrieval to the seed's variant, allowing the review queue to be dominated by positives from other patterns. Conversely, nearest-neighbor expansion uses the seed geometry but does not optimize attribute density, so it returns the same pattern at mostly ordinary, low-positive-rate items.

\begin{figure}[!t]
\centering
\includegraphics[width=0.93\textwidth]{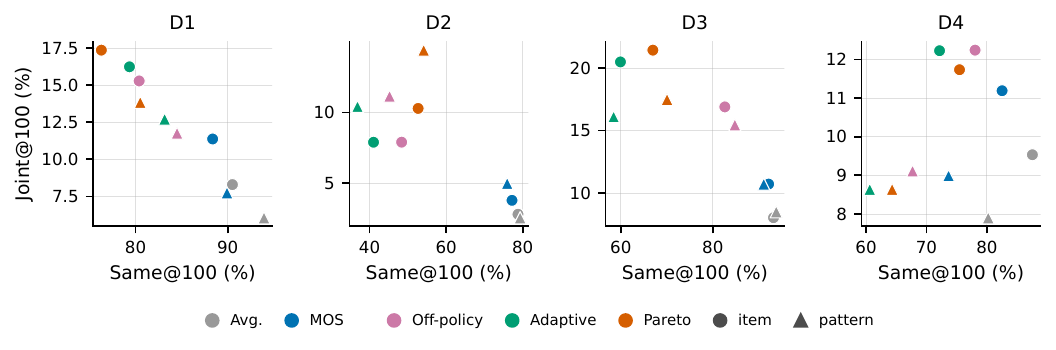}
\caption{The attribute--pattern tradeoff, per domain. Each point places a model at its same-pattern share (\metricSame{}, $x$) and intersection density (\shortJoint{}, $y$) on the item ($\circ$) and pattern ($\triangle$) holdouts. Average pooling and \mos{} (SFT) keep high \metricSame{} but lower \shortJoint{}. Off-policy is standardized at the common $100$-step endpoint, while Adaptive and Pareto use disjoint train-side validation. Pareto raises \shortJoint{} over both preference controls on D1--D3; its final-policy \metricSame{} effect is domain-dependent, increasing over Adaptive on D2/D3 but decreasing on D1. D4 is the boundary where Pareto provides no additional Joint gain.}
\label{fig:pareto}
\end{figure}

The task therefore sits between two standard tools: dense approximate nearest-neighbor retrieval \citep{karpukhin2020dense,johnson2019billion} and relevance-feedback query construction \citep{rocchio1971relevance}. Averaging the seed embeddings stays close to the seed pattern but retrieves mostly ordinary examples; optimizing only for the target attribute retrieves many positives from the wrong patterns. The matched evaluation bears out both sides of this tradeoff. AvgPool attains the highest \metricSame{} in all eight domain--split settings ($78.75$--$93.94\%$) but only $2.56$--$9.53\%$ \shortJoint{} (Table~\ref{tab:main}). Conversely, unconstrained attribute-oriented optimization improves \shortJoint{} by sacrificing pattern preservation, and continued optimization raises raw attribute density as \metricSame{} falls (Figures~\ref{fig:pareto} and \ref{fig:drift}). A matched D1 diagnostic that varies only the 50 embeddings averaged into the ANN query directly isolates these one-sided failures: a seed-independent Attribute-oriented query loses Pattern, a Pattern-only query remains lower-Attribute, and the within-Pattern high-density query achieves the highest \shortJoint{} on both holdouts (Appendix~\ref{app:three-trigger}). The desired query moves from the seed pattern toward a nearby high-density region of the same pattern.

We study this problem in continuous generative retrieval, using the term descriptively for methods that directly generate a continuous representation consumed by nearest-neighbor search. This formulation follows prior diffusion-based work in cross-modal retrieval, where CompoDiff synthesizes a retrieval feature in CLIP space, and in sequential recommendation, where DreamRec generates an oracle item embedding from a behavior history \citep{gu2024compodiff,yang2023dreamrec}. Each item in our setting has a frozen multimodal embedding. A generative retriever receives a sequence of item embeddings and produces a query embedding; the query then retrieves the top-$K$ items whose embeddings are most similar to it, by approximate nearest-neighbor (ANN) search over a large vector index. Continuous generation is attractive because the output query is not restricted to an existing item or to a fixed average of inputs. However, it raises a central supervision question: what target teaches the model to move toward higher-density examples while preserving the pattern?

We propose \textbf{\method{}}: \morphfullname{}, a staged training framework for attribute-seeking and pattern-preserving generative retrieval; Figure~\ref{fig:task} illustrates the task and the staged pipeline. The first stage is large-scale raw-sequence pretraining, which gives the diffusion transformer a general continuous retrieval prior. The next two stages form Metric-Ordered Sequence (\mos{}) training. The second stage of the full pipeline is \mos{} continuation pretraining (CPT), shared across domains. We use online recall density as an accessible proxy metric: for an anchor embedding, attribute density is the fraction of top-$K$ retrieved neighbors that satisfy the target attribute. Because dense online labeling is expensive, we label a sparse subset online, train a lightweight metric predictor over frozen embeddings, and use the predictor to order items inside each latent pattern cluster. The resulting training examples are sequences that move from lower-metric to higher-metric items within the same pattern; mixing the ordered sequences of all domains lets one continuation-pretrained model internalize the metric-improvement direction as a general capability rather than a per-domain memorization.

The third stage is \mos{} tail-centroid supervised fine-tuning (SFT). It maps the input prefix of a metric-ordered sequence to the centroid of its high-metric tail. The centroid target aggregates multiple high-metric examples, reducing sensitivity to a single noisy top item. The fourth stage is \hppo{}, which labels candidate query embeddings with the final online intersection metric and applies iterated reference-anchored preference optimization.\ Within \hppo{}, a tradeoff-aware Pareto pair filter removes winner pairs that reduce pattern purity, excluding the simplest training-time exchange of same-pattern share for higher intersection reward. The filter improves the evaluated attribute--pattern tradeoff on D1--D3; the D4 boundary case is documented in Appendix~\ref{app:dbexception}. This mirrors the role of post-training alignment: CPT and SFT provide a stable pattern-preserving generator, while \hppo{} aligns the final query distribution with the true online objective.

Our contributions are:
\begin{itemize}
    \item We formalize \textbf{pattern-preserving attribute retrieval} in this continuous-query setting, define an intersection metric (\shortJoint{}) that penalizes both attribute failure and pattern drift, and evaluate it under strict item- and pattern-holdout protocols.
    \item We introduce \textbf{\mos{} construction}, which turns sparse online retrieval labels into large-scale in-pattern training trajectories that move from low to high attribute density within a single pattern.
    \item We show that \textbf{\mos{} CPT/SFT} ---shared multi-domain continuation pretraining followed by domain-specific tail-centroid supervised fine-tuning---improves same-pattern attribute retrieval across four large-scale domains.
    \item We introduce \textbf{\hppo{}}, which adapts reference-anchored pairwise preference optimization to a continuous diffusion retriever using a flow-matching preference surrogate and true online \shortJoint{} labels over complementary static and policy-generated candidates. Its complete variant regenerates candidates from the current policy and applies a tradeoff-aware Pareto pair filter; it improves \shortJoint{} over SFT on seven of eight domain--split cells.
\end{itemize}

\section{Related Work}

\textbf{Generative retrieval.}
Generative retrieval reframes retrieval as a generation problem, producing document identifiers, semantic identifiers, or query representations rather than directly scoring all items; representative examples include autoregressive entity retrieval, differentiable search indices, and semantic-ID recommenders \citep{decao2021autoregressive,tay2022differentiable,rajput2023recommender}. These methods decode discrete identifiers, which is inherently serial and can hallucinate code sequences that match no real item, since most code combinations are invalid. An adjacent generate-then-encode route is HyDE, which generates a hypothetical document and re-encodes it for dense text retrieval \citep{gao2023hyde}. Direct continuous generation instead synthesizes the retrieval representation itself: CompoDiff generates a feature for cross-modal nearest-neighbor retrieval \citep{gu2024compodiff}, while DreamRec generates an oracle next-item embedding from a behavior history \citep{yang2023dreamrec}. Subsequent work extends this line through spherical next-interest generation followed by item retrieval in DimeRec \citep{li2025dimerec}, one-step oracle-item generation with adaptive preference alignment in TA-Rec \citep{mao2025efficiency}, and continuous-token diffusion with latent representations for Top-$K$ retrieval in ContRec \citep{qu2026continuous}. X-Rec studies continuous generative retrieval for large-scale recommendation, using anchor conditioning, Riemannian flow matching, and a late-interaction diffusion transformer to generate embedding triggers for ANN retrieval \citep{shen2026xrec}. Our contribution changes the task and supervision: from a multimodal item sequence we generate a query embedding under a pattern-preserving attribute-density objective, returning items that are both attribute-positive and of the seed pattern rather than predicting the next item or retrieving within a single modality. Because the conditioning input is an item sequence encoded with self-attention, the encoder side relates to self-attentive sequential recommendation \citep{kang2018sasrec}, but our objective is pattern-preserving attribute retrieval through a generated query rather than next-item prediction. It also connects to classical ideas: unlike dense retrieval \citep{karpukhin2020dense,johnson2019billion} with fixed learned embeddings, we synthesize the query itself; unlike Rocchio relevance feedback \citep{rocchio1971relevance} with its fixed update rule, our tail-centroid target is a learned conditional mapping from the seed sequence; and \mos{} construction relates to curriculum learning \citep{bengio2009curriculum}, except the order encodes a retrieval direction---low to high attribute density within a pattern---rather than example difficulty.

\textbf{Dense pseudo-relevance feedback.}
ANCE-PRF refines a dense text-query representation by jointly encoding the query and the top passages from an initial ANCE retrieval, while keeping the passage index fixed for a second retrieval \citep{yu2021improving}. We adapt this two-pass mechanism to the frozen item-embedding space: a dedicated FirstP query encoder is trained on the full strict D1 pool with two ANN hard-negative refreshes, and the inherited PRF encoder consumes the first three unlabeled FirstP results. A matched $k=0$ branch isolates the effect of feedback. Table~\ref{tab:appdc} in Appendix~\ref{app:perdomain} reports this task adaptation under the same D1 holdouts and deterministic cases as our existing baselines.

\textbf{Continuous generative models.}
Continuous generation synthesizes a target vector directly rather than decoding discrete tokens, avoiding the quantization and codebook design that discrete generative retrievers require. Diffusion and score-based models \citep{ho2020denoising,song2021score} learn to reverse a noising process, and their flow-matching / rectified-flow formulations \citep{lipman2023flow,liu2023flow} recast this as learning a velocity field that transports noise to data along nearly straight trajectories, so generation reduces to integrating an ordinary differential equation in a few steps. The diffusion transformer replaces the convolutional U-Net backbone with a transformer and injects the diffusion timestep through adaptive layer normalization \citep{peebles2023scalable}, which scales well and fits naturally when the conditioning signal is itself a sequence; classifier-free guidance \citep{ho2022classifierfree} then trades sample fidelity against conditioning strength at inference time. Most of this literature generates perceptual signals---images, video, or audio---but a related line shows that continuous generation can replace discrete token prediction inside autoregressive models by modeling each continuous token with a small diffusion head \citep{li2024autoregressive}. We adopt a diffusion transformer with a flow-matching objective but generate in a frozen item-embedding space: the output is a query embedding consumed by nearest-neighbor retrieval rather than a signal to be rendered. Diffusion has also been applied to recommendation by denoising user interaction vectors \citep{wang2023diffrec}; we instead generate a query embedding for downstream retrieval rather than reconstructing interactions.

\textbf{Preference optimization and alignment.}
Reinforcement learning from human feedback aligns generative models with preference signals through a learned reward model \citep{ouyang2022training}. Direct Preference Optimization removes that reward model and trains directly on preference pairs with a reference-anchored classification loss \citep{rafailov2023direct}, and the same objective has been adapted to diffusion models \citep{wallace2024diffusion}. A parallel line tunes generative models with policy-gradient RL: denoising is cast as a multi-step decision process \citep{black2023training}, and GRPO \citep{shao2024deepseekmath} has recently reached flow-matching generation by converting the deterministic sampling ODE into an SDE with tractable per-step log-probabilities \citep{liu2025flowgrpo,xue2025dancegrpo}. We take the preference route rather than on-policy GRPO because our reward is real online retrieval through a rate-limited service, where labeling candidates once offline is far cheaper than an on-policy sampling loop. \hppo{} adapts preference optimization to continuous query generation in three ways: a candidate's implicit score is its negative flow-matching loss rather than a token log-probability, preference labels come from real online retrieval rather than human raters or a learned reward model, and the candidate pool mixes deterministic constructions with policy samples (complementary in our ablation). It further adds a tradeoff-aware Pareto pair filter that keeps preference pairs non-decreasing in a secondary objective (pattern purity), removing pair-level updates that improve the primary reward by directly sacrificing purity and thereby resisting the reward over-optimization an unconstrained proxy invites \citep{gao2023overoptimization}. This relates to multi-objective alignment such as multi-objective DPO, which scalarizes several reward models into one preference objective \citep{zhou2024modpo}; our filter instead enforces the secondary objective at the pair-construction level, with no additional reward models.

\section{Preliminaries}
\label{sec:background}

\paragraph{Items and embeddings.}
Each item is a piece of multimodal content (for example, video, image, text, video--text, or image--text). A frozen pretrained multimodal embedding model maps each item $i$ to an embedding $e_i\in\mathbb{R}^{d}$; these embeddings are fixed throughout and define the retrieval space. A generative retriever receives a sequence of item embeddings and produces one or more query embeddings,
\begin{equation}
    q=f_\theta(e_{i_1},\ldots,e_{i_m}),\qquad q\in\mathbb{R}^{d},
\end{equation}
each searched in the same vector index as the item embeddings.

\paragraph{Attributes and patterns.}
Each domain $y$ exposes two signals over items. A pre-existing production \emph{attribute-scoring system} gives a binary label $a_y(i)\in\{0,1\}$ for whether item $i$ meets a target attribute---a quality, safety, policy, originality, or integrity criterion. We use this output as the domain's fixed operational label. The scorer is coarse: it does not separate the many distinct sub-types that exist within one attribute. To recover finer structure we fit $C$ $k$-means centers $\{\mu_{y,r}\}_{r=1}^{C}$ on the attribute-positive items in embedding space. The positive pool determines only these centers: every item in the retrieval index---including those with $a_y(i)=0$---receives the nearest-center proxy assignment
\begin{equation}
    c_y(i)=\arg\min_{r\in\{1,\ldots,C\}}\lVert e_i-\mu_{y,r}\rVert_2^2.
\end{equation}
We use $c_y(i)$ as a \emph{pattern} proxy---a fine-grained latent neighborhood within the attribute, rather than a human-labeled class. Human labeling of every item at this granularity is prohibitively expensive, so embedding clustering provides a scalable, label-free approximation.

We quantitatively validate this frozen proxy on a human-annotated pattern evaluation set whose labels were produced independently of the proxy construction and were not derived from its cluster assignments (Appendix~\ref{app:patternproxy}). The set contains $66$ human patterns, $15{,}754$ pattern--item memberships, and $15{,}414$ unique annotated items. Because the proxy clusters are substantially finer-grained than the human patterns, we test enrichment rather than exact one-to-one correspondence. Same-human-pattern pairs map to the same proxy cluster at $18.19\%$, versus $2.87\%$ under a matched permutation ($6.33\times$, $p<0.0005$); conversely, $80.00\%$ of same-cluster annotated pairs share a human pattern, versus $10.89\%$ for matched random pairs ($7.34\times$). These results show strong non-random alignment with human annotations while retaining the operational interpretation of clusters as latent proxies. The attribute scorer and pattern proxy support offline data construction, evaluation metrics, and training rewards. The scorer is an online production model and the clustering is computed offline; neither is invoked by the generative retriever to construct a query at inference time.

\paragraph{Attribute-scorer validation.} The production scoring systems used to construct the D1--D4 labels are evaluated through an established offline quality-assurance process on finalized, versioned test sets aligned with their binary operational criteria. Each scorer achieves at least $80\%$ precision and $80\%$ recall on its corresponding test set. The evaluation sets combine a human-reviewed, score-stratified core, used to recover rare positives, with prevalence-calibrated samples from the broader content pool; selected high-risk model-disagreement cases receive additional review before finalization. This validation supports using the scorer outputs as fixed binary operational labels in our experiments. The fine-grained pattern proxy is defined and evaluated separately. The lightweight metric predictor used during \mos{} construction is also distinct: it orders training items, whereas final retrieval metrics evaluate retrieved items with the fixed production scorer.

\paragraph{Cluster-count selection.} We select the number of pattern clusters from training-side semantic-alignment and construction-support diagnostics rather than held-out retrieval performance. Appendix~\ref{app:c-sensitivity} reports a controlled D1 sweep over $C\in\{500,1000,2000\}$ and identifies $C=1000$ as the observed granularity elbow between merging and fragmentation; this is a domain-specific deployment choice rather than a universal cluster count.

\paragraph{Task.}
A seed sequence $X=(e_{i_1},\ldots,e_{i_m})$ is drawn from a single pattern $c$, so all seeds share the pattern. The goal is to generate a query whose top-$K$ retrieved neighbors are dense in items that are \emph{both} attribute-positive \emph{and} of the seed pattern. Averaging the seed embeddings stays inside the pattern but lands in a low-attribute region, while optimizing only for the attribute drifts to other patterns; the desired query therefore \emph{seeks} higher attribute density while \emph{preserving} the seed pattern.

\paragraph{Metrics.}
All metrics are densities over the top-$K$ retrieved set $\ret_K(q)$, i.e.\ properties of the retrieved neighborhood rather than of a single item. The raw attribute density
\begin{equation}
    D_y(q)=\frac{1}{K}\sum_{j\in \ret_K(q)} a_y(j),
\end{equation}
reported as \metricIssue{}, is insufficient alone: it rewards attribute hits even when the query drifts to a different pattern. The primary metric is the intersection density
\begin{equation}
    J_y(q,c)=\frac{1}{K}\sum_{j\in \ret_K(q)}
    a_y(j)\ind[c_y(j)=c],
\end{equation}
reported as \metricJoint{}, which requires both attribute correctness and pattern preservation. We also report the conditional same-pattern purity among retrieved attribute positives,
\begin{equation}
    P_y(q,c)=
    \frac{\sum_{j\in \ret_K(q)} a_y(j)\ind[c_y(j)=c]}
    {\max(\sum_{j\in \ret_K(q)} a_y(j),1)},
\end{equation}
reported as \metricCond{}. The denominator guard assigns a placeholder value of zero when no attribute positives are retrieved; such queries are excluded from the reported \metricCond{} mean. We also report the same-pattern share
\begin{equation}
    S_y(q,c)=\frac{1}{K}\sum_{j\in \ret_K(q)} \ind[c_y(j)=c],
\end{equation}
reported as \metricSame{}, the fraction of retrieved items assigned to the seed pattern regardless of the attribute. Because the nearest-center assignment is defined for every retrieved item, \metricSame{} includes both attribute-positive and attribute-negative items; \metricJoint{} additionally requires $a_y(j)=1$.

\section{Method}

\method{} combines Metric-Ordered Sequence (\mos{}) training and Hybrid-Policy Preference Optimization (\hppo{}) on a pretrained diffusion-transformer backbone (Figure~\ref{fig:task}). The complete pipeline has four weight-updating stages: (i) large-scale raw-sequence pretraining, (ii) shared multi-domain \mos{} continuation pretraining (CPT), (iii) domain-specific \mos{} tail-centroid supervised fine-tuning (SFT), and (iv) \hppo{} alignment. \mos{} construction (Section~\ref{sec:mos-construction}) produces the in-pattern sequences used by \mos{} CPT/SFT (Section~\ref{sec:mos}); \hppo{} then aligns the resulting generator with online retrieval metrics (Section~\ref{sec:hppo}). All four training stages share one diffusion-transformer backbone. Raw-sequence pretraining and \mos{} CPT/SFT use flow-matching supervision with different data or targets; \hppo{} instead uses flow-matching losses to construct a reference-anchored preference objective with a tail-centroid anchor.

\subsection{Continuous Generative Retriever}

\begin{figure}[!t]
\centering
\includegraphics[width=1\textwidth]{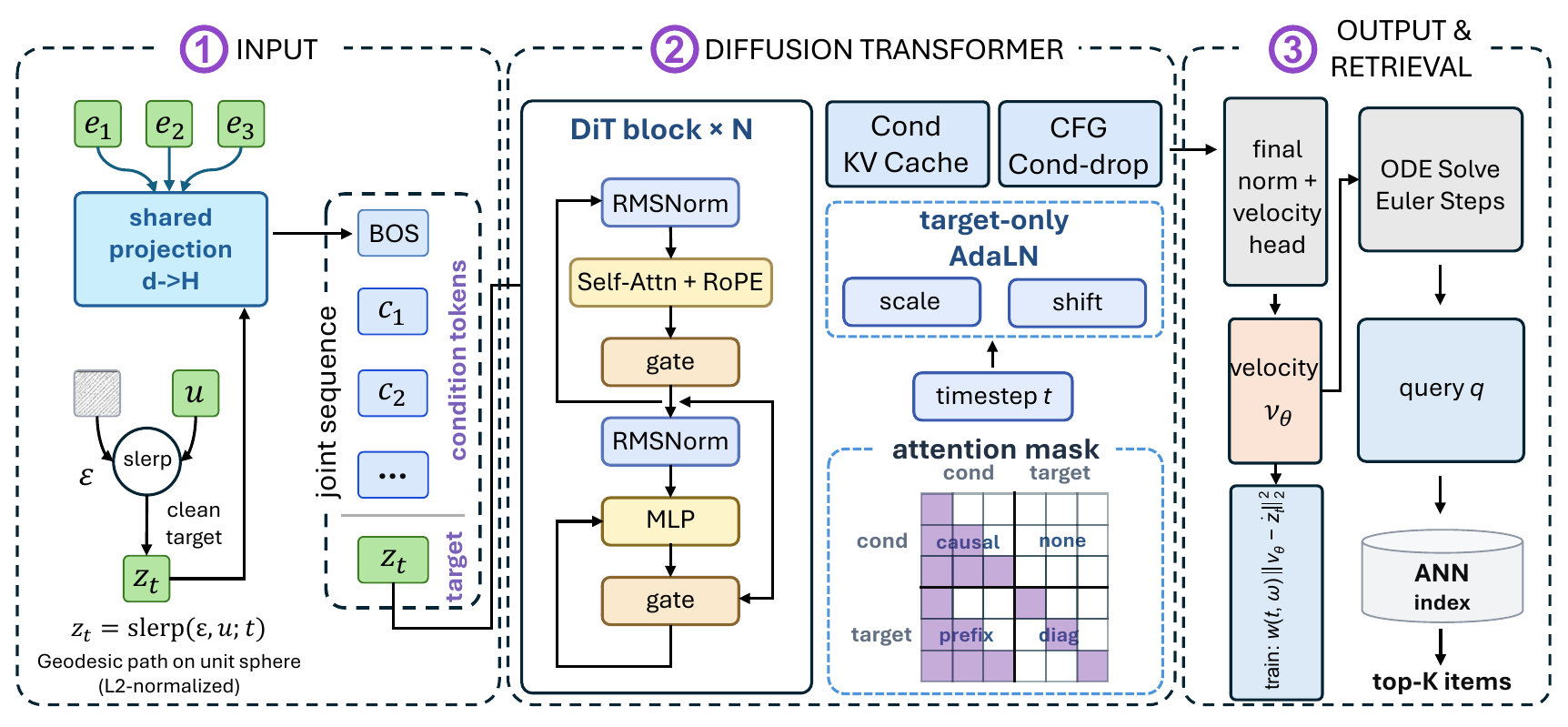}
\caption{\textbf{Continuous generative retriever architecture.} Frozen item embeddings are projected into condition tokens and concatenated with a noised target token on a spherical flow-matching path. A single diffusion transformer jointly processes condition and target tokens under a structured mask: condition tokens encode the prefix, each target attends to its condition prefix but not to other targets, and condition positions do not attend to targets. The target timestep is injected through AdaLN, the velocity head predicts the flow field, and ODE integration produces a $d$-dimensional query embedding that is searched directly in the ANN index.}
\label{fig:arch}
\end{figure}

Figure~\ref{fig:arch} sketches the architecture; the retriever operates entirely in the frozen item-embedding space defined in Section~\ref{sec:background}.

The model is a single diffusion-transformer backbone \citep{vaswani2017attention,peebles2023scalable} trained with a flow-matching objective (defined in the \mos{} training subsection, Section~\ref{sec:mos}): the architecture is the diffusion transformer, and the generative objective is flow matching, as in recent continuous generators. It processes the condition tokens and the noised target tokens jointly, rather than compressing the condition into one vector and feeding it to a separate generation head. Each condition position projects its item embedding to the transformer width and adds a positional encoding and a learned role marker that separates condition positions from target positions; a binary condition mask marks valid positions so batched sequences can use different effective prompt lengths.

Each target position holds a noised target token, a point on the flow-matching path between Gaussian noise projected onto the unit sphere and the normalized clean target at a sampled time $t$ (the objective is defined in that subsection). The condition tokens and the noised target token(s) are concatenated into one sequence and passed through the transformer under a structured attention mask: condition positions attend causally among themselves; each target attends to its condition prefix but not to other targets, so several targets are generated independently in one pass; and condition positions do not attend to targets, so the condition is encoded once and reused. The time $t$ is injected at the target positions through adaptive layer normalization, and the network predicts a flow-matching velocity; integrating the resulting ODE from unit-normalized Gaussian noise yields the output query embedding in the same $d$-dimensional space as the item embeddings, which is searched directly in the vector index with no quantization or identifier decoding. Because each target attends to the full condition sequence rather than a pooled summary, the model retains fine-grained information from the seed items; because the output is continuous, it can synthesize a query that lies between observed items rather than copying one.

\subsection{\mos{} Construction}
\label{sec:mos-construction}

\mos{} construction aims to obtain training targets that move a query in the desired direction. For each domain $y$, we first collect items that are positive under the domain scorer and cluster their frozen embeddings into $C$ groups with $k$-means. These clusters are not assumed to be human-labeled semantic classes. They are scalable latent pattern proxies: they define the local region in which the generated query should remain, and they allow evaluation to distinguish same-pattern positives from arbitrary positives.

Next, we estimate where high-density regions lie inside each pattern. For a sampled set of anchor embeddings, we compute online recall density by retrieving top-$K$ neighbors from the production-like vector index and applying the target scorer to the recalled items. This label measures the actual retrieval consequence of using the anchor as a query. It is expensive because it requires retrieval, item lookup, and scorer access; therefore, \method{} labels only a subset online.

The sparse labels are expanded to the full item pool with a density predictor. We train a ridge regression predictor $\hat{D}_y(e)$ from frozen embeddings to online recall density. The predictor is deliberately low-capacity because \mos{} construction needs only a useful ordinal signal, not a calibrated density estimate or an expressive final retriever. Across the available strict pattern-out and super-pattern-out cells, it retains AUC20/80 of $0.882$--$0.973$ and Spearman correlation of $0.530$--$0.752$ (Appendix Table~\ref{tab:density}); ascending predicted-density order then outperforms random and descending order in all eight item- and pattern-holdout domain cells (Table~\ref{tab:order}). A predicted value is never reported as a retrieval result: it only determines the relative order of items during \mos{} construction, while all final metrics are computed after real top-$K$ vector retrieval.

\begin{figure}[!t]
\centering
\includegraphics[width=0.8\textwidth]{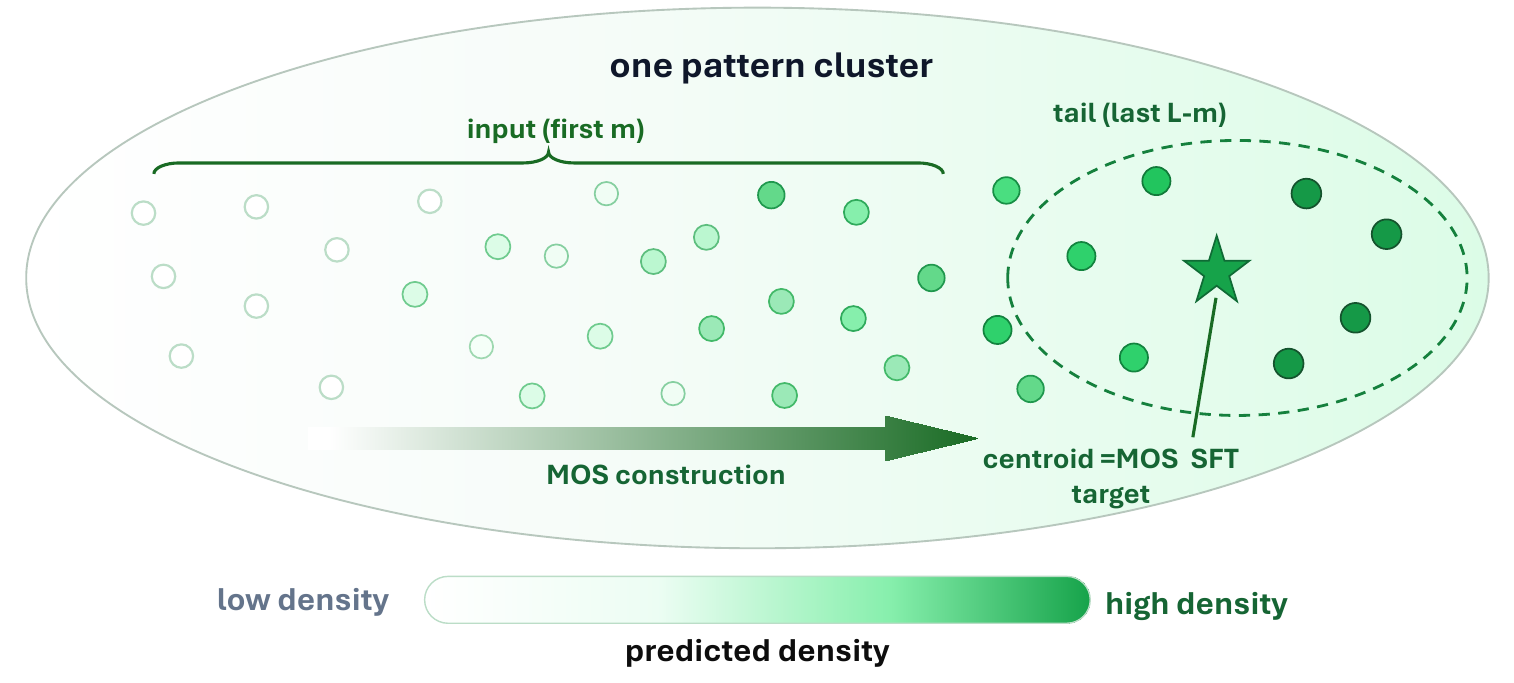}
\caption{\textbf{\mos{} construction of in-pattern trajectories.} Within each latent pattern cluster, items are scored by a lightweight predictor of online recall density and sorted from low to high predicted attribute density. The predictor is used only to order training sequences; final metrics are always computed by real top-$K$ retrieval. Continuation pretraining supervises local prefix-to-target transitions along the ordered trajectory, while tail-centroid SFT maps the first $m$ items to the centroid of the final $L-m$ high-density tail. This converts sparse online labels into dense same-pattern supervision for learning the direction from ordinary in-pattern items toward high-attribute-density regions.}
\label{fig:traj}
\end{figure}

Within each latent pattern cluster, we sample candidate groups and sort them by predicted density from low to high (Figure~\ref{fig:traj}); Algorithm~\ref{alg:construction} summarizes the construction. Each emitted sequence has a fixed length $L$. Normal items are assigned to the same domain-specific cluster space as positive items and are sampled only from the current cluster and density bin. They may be mixed at a fixed aggregate ratio, but the ratio is not forced within every sequence. If a group cannot be constructed, that attempt ends without drawing replacements from other clusters. Thus early and late positions remain within the same pattern while moving from lower to higher predicted density.

We validate the predictor under item-hash, pattern-out, and super-pattern-out splits (Appendix~\ref{app:patternproxy}). A pattern is one fine-grained cluster used by \mos{} construction and retrieval evaluation. For the super-pattern diagnostic, we average the $L_2$-normalized labeled-anchor embeddings within each fine cluster and group the resulting centroids with MiniBatchKMeans, producing coarser embedding-derived families. Pattern-out withholds individual fine clusters from predictor fitting, whereas super-pattern-out withholds entire families and thus removes geometrically related fine clusters together. These group-level holdouts test whether embedding geometry predicts density beyond a memorized mean for seen clusters.

\begin{algorithm}[!t]
\caption{\mos{} Construction}
\label{alg:construction}
\begin{algorithmic}[1]
\STATE Input: positive pool $\mathcal{P}_y$, optional domain-specific normal pool $\mathcal{N}_y$, clusters $c_y$, predictor $\hat{D}_y$
\FOR{cluster $c$}
    \STATE collect positives $\mathcal{P}_{y,c}=\{i\in\mathcal{P}_y:c_y(i)=c\}$
    \STATE collect normal items $\mathcal{N}_{y,c}=\{i\in\mathcal{N}_y:c_y(i)=c\}$
    \STATE sample candidate groups from $\mathcal{P}_{y,c}$ and $\mathcal{N}_{y,c}$ within the selected density bins; skip infeasible groups
    \STATE score each item with $\hat{D}_y(e_i)$
    \STATE sort each group by predicted density from low to high
    \STATE emit a length-$L$ sequence
\ENDFOR
\STATE Output: length-$L$ metric-ordered training sequences
\end{algorithmic}
\end{algorithm}

\subsection{\mos{} Training}
\label{sec:mos}

The generated query should both preserve the input pattern and move toward the high-metric tail of the ordered sequence. \mos{} training uses two stages to separate these requirements; they build on a backbone first pretrained on large-scale item sequences that carry a natural order (for example, session sequences, user behavior sequences, or time-ordered item streams), with the same flow-matching continuation objective---a raw-sequence pretraining that is not metric-ordered and only establishes a general continuous-retrieval prior.

The first stage of \mos{} training is continuation pretraining (CPT). Given a metric-ordered sequence $S=(i_1,\ldots,i_{L})$, the model is trained with dense prefix supervision: each target position predicts its own item embedding from the preceding prefix. This objective exposes the model to many local transitions along the low-to-high metric trajectory. It teaches the model the geometry of metric improvement before the final task asks it to produce one query embedding. Continuation pretraining is multi-domain: the ordered sequences of all domains are mixed into one shuffled pool, with per-domain repetition weights so that smaller domains are not drowned out, and a single shared model is trained over the pool. The subsequent stages are domain-specific: SFT initializes from the shared CPT checkpoint, and \hppo{} initializes from the resulting domain-specific SFT checkpoint.

We train the generator by flow matching \citep{lipman2023flow,liu2023flow}. Because the retrieval embeddings are $L_2$-normalized and therefore lie on the unit hypersphere, we use the spherical (Riemannian) instantiation of flow matching \citep{chen2024riemannian}. For a conditioning sequence $X$ and a target embedding $u$, the model first normalizes the target to $\bar{u}=u/\|u\|_2$ and samples unit-length noise $\epsilon=g/\|g\|_2$, where $g\sim\mathcal{N}(0,I)$. We sample a time $t$ from a logit-normal schedule~\citep{esser2024scaling} ($t=\sigma(\xi),\ \xi\sim\mathcal{N}(0,1)$, with $\sigma$ the logistic function), form the geodesic path $z_t$ from $z_0=\epsilon$ to $z_1=\bar{u}$, and regress the network velocity onto the path velocity $\dot{z}_t$ with a Jacobi curvature weight:
\begin{equation}
    \label{eq:fm}
    \mathcal{L}_{\mathrm{FM}}(X,u)
    =
    \mathbb{E}_{t,\epsilon}
    \left[
    w(t,\omega)
    \left\|
    v_\theta(z_t,t,X)-\dot{z}_t
    \right\|_2^2
    \right],
\end{equation}
where $v_\theta$ is the transformer velocity field and
\[
\begin{aligned}
    \omega &= \arccos\!\left(\operatorname{clip}
    \left(\epsilon^\top\bar{u},-1+10^{-7},1-10^{-7}\right)\right),\\
    w(t,\omega) &= \left(\frac{\sin(t\omega)}{t\omega}\right)^2 .
\end{aligned}
\]
Time advances from noise to target; the ratio is evaluated as $1$ when $t\omega$ is numerically near zero, consistent with its limit. At inference we integrate $\dot{z}=v_\theta$ from $z_0=\epsilon$ to produce the query embedding. We use this weighted per-target loss for \mos{} CPT/SFT and \hppo{}. For a metric-ordered sequence $S=(i_1,\ldots,i_{L})$, continuation pretraining applies it to supervised prefix-target pairs:
\begin{equation}
    \mathcal{L}_{\mathrm{CPT}}
    =
    \sum_{r\in\mathcal{I}}
    \mathcal{L}_{\mathrm{FM}}
    \left((e_{i_1},\ldots,e_{i_{r-1}}), e_{i_r}\right),
\end{equation}
where $\mathcal{I}$ is the set of supervised target positions (dense supervision covers every position with a preceding prefix), each predicting its item $e_{i_r}$ from that prefix $e_{i_1},\ldots,e_{i_{r-1}}$; the diagonal target--target mask lets all of them be supervised in a single pass. This objective gives the model dense training signal along the low-to-high metric trajectory.

The second stage of \mos{} training is tail-centroid supervised fine-tuning (SFT). Each ordered sequence is converted into one input-target pair. The input is the first $m$ embeddings:
\begin{equation}
    X=(e_{i_1},\ldots,e_{i_{m}}).
\end{equation}
The target is the centroid of the final $L-m$ embeddings:
\begin{equation}
    t_{\mathrm{tail}}=\frac{1}{L-m}\sum_{r=m+1}^{L}e_{i_r}.
\end{equation}
Before constructing the spherical path, the model normalizes the tail centroid to $t_{\mathrm{tail}}/\|t_{\mathrm{tail}}\|_2$ through the target-normalization step in $\mathcal{L}_{\mathrm{FM}}$. The SFT loss is
\begin{equation}
    \mathcal{L}_{\mathrm{SFT}}
    =
    \mathcal{L}_{\mathrm{FM}}
    \left((e_{i_1},\ldots,e_{i_{m}}), t_{\mathrm{tail}}\right).
\end{equation}
The centroid target is a stable summary of multiple high-metric same-pattern examples. It is less noisy than a single top item and less narrow than a small top-$k$ target. This is also the reason we do not supervise the model to exactly reproduce the highest-metric item: a top item may be an outlier, while the tail centroid represents a local high-metric region.

In short, each stage changes one component from the previous. Pretraining establishes the prior with a prefix-continuation objective (the per-position $\mathcal{L}_{\mathrm{FM}}$) on raw, non-metric-ordered sequences; CPT keeps that objective but switches the data to metric-ordered sequences; and SFT keeps the metric-ordered data but switches the supervision target to the single tail centroid $t_{\mathrm{tail}}$. The preference stage then uses online-labeled candidate comparisons as its main learning signal, while retaining the SFT tail centroid as an auxiliary anchor.

\subsection{Hybrid-Policy Preference Optimization (\hppo{})}
\label{sec:hppo}

\begin{figure}[!t]
\centering
\includegraphics[width=1\textwidth]{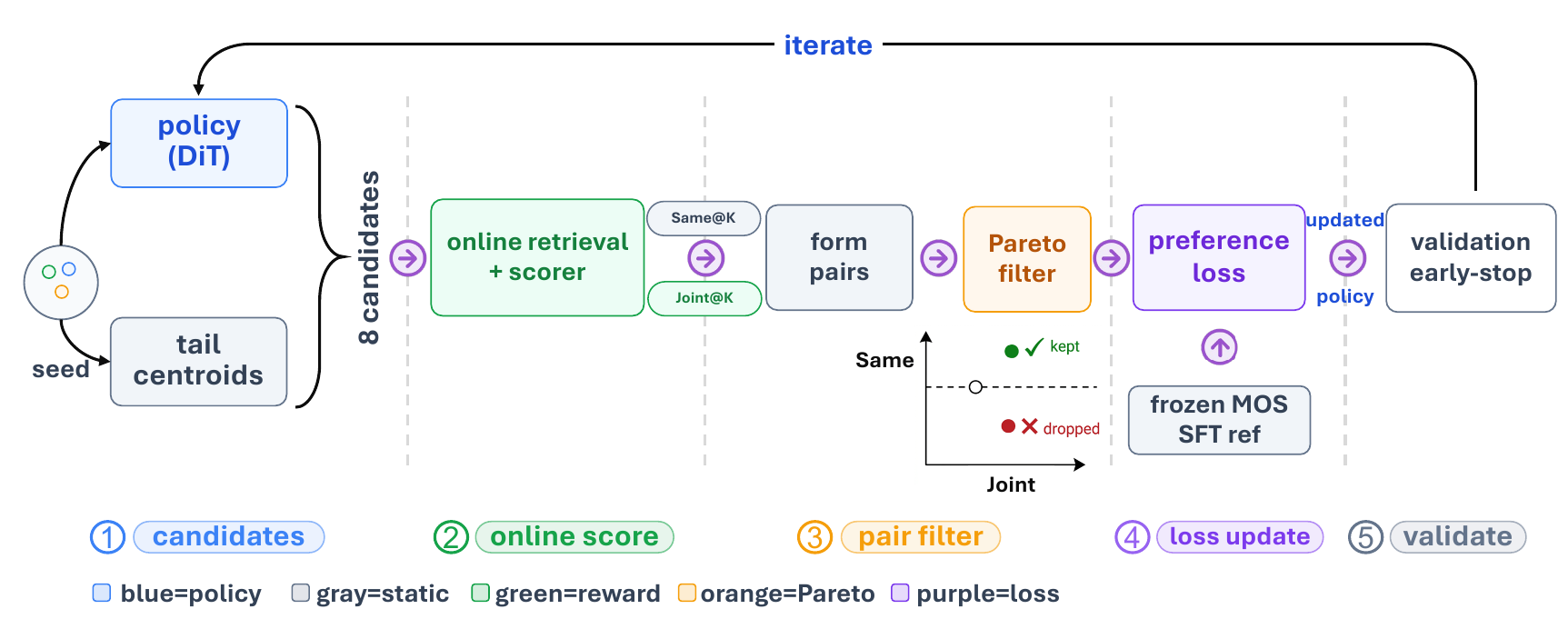}
\caption{\textbf{Complete Pareto-filtered Hybrid-Policy Preference Optimization.} For each seed case, HPPO builds a hybrid candidate pool from deterministic tail-centroid constructions and policy-generated guidance-fan queries, labels every candidate with real online \shortJoint{} and \shortSame{}, and forms preference pairs from candidates separated by a reward margin. The Pareto pair filter keeps a winner--loser pair only when the higher-Joint winner does not reduce same-pattern share relative to the loser, removing that pair-level shortcut from training. A frozen SFT reference and tail-anchor loss regularize the reference-anchored preference update, while on-policy rounds regenerate the policy candidates and select checkpoints by disjoint train-side online validation.}
\label{fig:hppo}
\end{figure}

\mos{} CPT/SFT uses constructed targets ordered by the proxy metric, not the final online intersection metric itself. The last stage, Hybrid-Policy Preference Optimization (\hppo{}), closes this gap by directly optimizing online \shortJoint{}. Figure~\ref{fig:hppo} sketches the stage.

For each training case, \hppo{} forms a candidate set of query embeddings from a hybrid policy. One part is deterministic and metric-based: tail top-$k$ centroids of the ordered sequence (averaging the top-$k$ items for each depth $k$ in a small set $\mathcal{K}$ and $L_2$-normalizing the result), which are strong pattern-preserving constructions. The other part is policy-generated: queries sampled from the current generative retriever. Because point-target tail-centroid SFT makes the conditional generator nearly deterministic, independently sampled policy queries collapse to almost the same embedding; we therefore diversify the policy candidates with a \emph{guidance-scale fan}, generating the policy part at several classifier-free guidance scales (a small set $\mathcal{S}$). The fan varies conditioning strength to produce candidates with different attribute--pattern tradeoffs. Its effect on attribute density and pattern purity depends on the domain and checkpoint (Appendix~\ref{app:bon-analyses}). Every candidate is searched in the real vector index and labeled by the primary intersection metric $J_y(q,c)$, so the preference signal is the true online objective rather than a learned proxy reward.

We retain a case only if its top-1/top-2 candidate reward gap is at least $\delta$. Within a retained case, the highest-reward candidate is $q^+$, and each candidate $q^-$ with $J_y(q^+,c)-J_y(q^-,c)\ge\delta$ is a potential loser, subject to the Pareto filter below. A forward pass samples $(t,\epsilon)$ and scores each candidate with the per-dimension negative flow-matching loss,
\begin{equation}
    \widehat{s}_\theta(X,q;t,\epsilon)
    =
    -\frac{1}{d}\widehat{\mathcal{L}}_{\mathrm{FM}}(X,q;t,\epsilon).
\end{equation}
Here $\widehat{\mathcal{L}}_{\mathrm{FM}}$ is one Monte Carlo draw of the Jacobi-weighted loss in Equation~\eqref{eq:fm}. Both policy and frozen-reference scores apply the same weighting formula and divide by the embedding dimension ($d=1536$ for D1--D4). This stochastic score is neither an exact nor a calibrated policy log-likelihood. We use it only as a \emph{reference-anchored relative preference surrogate}, in the spirit of Diffusion-DPO \citep{wallace2024diffusion}: during training, $s_\theta$ below denotes one Monte Carlo draw of $\widehat{s}_\theta$, and a frozen SFT reference prevents unconstrained drift. Table~\ref{tab:surrogate-main} directly evaluates when this relative surrogate agrees with the online-reward ordering and how its Monte Carlo stability changes with reward separation.
\begin{equation}
    \mathcal{L}_{\mathrm{pref}}
    =
    -\log \sigma
    \left(
    \beta
    \left[
    s_\theta(X,q^+)-s_\theta(X,q^-)
    -
    s_{\mathrm{ref}}(X,q^+)+s_{\mathrm{ref}}(X,q^-)
    \right]
    \right).
\end{equation}
Here $\beta$ is the preference temperature. The \mos{} CPT/SFT stages thus provide the stable pattern-preserving initialization, and \hppo{} aligns the final query distribution with the online objective.

\paragraph{Tradeoff-aware pair filter.} Optimizing the intersection metric naively still admits a degenerate shortcut: a winner can earn a higher $J_y$ by retrieving many more attribute positives while drifting to other patterns, as long as the raw attribute gain outweighs the purity loss. Preferring such winners pushes the policy toward global attribute retrieval, exactly the failure mode the task is meant to avoid. We therefore add a Pareto-dominance constraint to pair construction using the same-pattern share $S_y(q,c)$ (\metricSame{}) defined in the problem setup. Among the margin-passing pairs, we keep the pair $(q^+,q^-)$ for the preference loss only if
\begin{equation}
    S_y(q^+,c)\ \ge\ S_y(q^-,c)-\varepsilon ,
\end{equation}
that is, the higher-reward winner must not lose same-pattern share relative to the loser (up to a tolerance $\varepsilon$, with $\varepsilon=0$ the strict setting). The filter is free: $S_y$ is already computed during online labeling, so no extra retrieval is needed. It removes purity-sacrificing pairs from the gradient and constrains the search against direct Same-for-Joint exchanges at pair construction; Table~\ref{tab:hppo-abl} measures the resulting final-policy tradeoff rather than assuming that the pairwise constraint guarantees a global increase in \metricSame{}.

\paragraph{Full objective.} A small anchor loss keeps preference updates near the pattern-preserving SFT solution by reusing its tail-centroid target $t_{\mathrm{tail}}$,
\begin{equation}
    \mathcal{L}_{\mathrm{anchor}} = \mathcal{L}_{\mathrm{FM}}(X, t_{\mathrm{tail}}),
\end{equation}
which uses the same Jacobi-weighted per-target loss without the $1/d$ score normalization. The top-50 tail mean is normalized when the HPPO anchor is constructed and passes through the model's target-normalization step before spherical interpolation. The full \hppo{} objective averages the preference loss over the retained pairs and adds this anchor,
\begin{equation}
    \mathcal{L} = \frac{1}{|\mathcal{P}|}\sum_{(q^+,q^-)\in\mathcal{P}} \mathcal{L}_{\mathrm{pref}}(q^+,q^-) \;+\; \lambda\,\mathcal{L}_{\mathrm{anchor}},
\end{equation}
where $\mathcal{P}$ contains the highest-reward-winner pairs from cases passing the top-1/top-2 gate, further restricted by the reward margin $\delta$ and the Pareto filter; $\lambda$ is the anchor weight.

\paragraph{Iterated hybrid policy with online validation.} The complete \hppo{} procedure uses an iterated on-policy hybrid candidate pool: in each round, the guidance-fan policy candidates are regenerated from the \emph{current} policy and relabeled online, while the static candidates and the frozen SFT reference stay fixed. To avoid the drift that uncontrolled iteration produces, each round is short (about one epoch of the kept pairs), the reference is always the original SFT model, and round selection uses a disjoint train-side validation slice: the initial best checkpoint is SFT, with its true online \shortJoint{} on this slice evaluated at the fixed serving guidance scale. Each round selects the best of its saved checkpoints on the same validation slice and accepts it only if its score strictly exceeds the best retained score. Otherwise training stops and returns the retained checkpoint, which is SFT if the first round fails to improve. For mechanism ablation, we also report \emph{off-policy}, which collects candidates once from the frozen SFT policy plus the static constructions, and \emph{adaptive}, which retains iteration but removes the pair filter. Algorithm~\ref{alg:hppo} summarizes the complete iterated Pareto-filtered procedure.

\begin{algorithm}[!t]
\caption{Iterated Pareto-Filtered \hppo{}}
\label{alg:hppo}
\begin{algorithmic}[1]
\STATE Input: SFT policy $\theta_0$; frozen reference $\theta_{\mathrm{ref}}\!=\!\theta_0$; cases $\{(X_n,c_n)\}$; centroid depths $\mathcal{K}$; guidance scales $\mathcal{S}$; margin $\delta$; tolerance $\varepsilon$; anchor weight $\lambda$; max rounds $R$; validation slice $\mathcal{V}$
\STATE $\theta \leftarrow \theta_0$;\quad $\theta^\star \leftarrow \theta_0$
\STATE $J^\star \leftarrow \operatorname{Validate}(\theta_0;\mathcal{V})$ using true online $J_y$ at the serving guidance scale
\STATE build static candidates (top-$k$ centroids, $k\in\mathcal{K}$) for every case and label them once by online retrieval, giving $(J_y,S_y)$
\FOR{round $r=1,\dots,R$}
    \STATE $\mathcal{P} \leftarrow \emptyset$
    \FOR{each case $n$}
        \STATE generate policy candidates from $\theta$ at each scale $s\in\mathcal{S}$ and label them online, giving $(J_y,S_y)$
        \STATE $\mathcal{C} \leftarrow \text{static}\ \cup\ \text{policy}$ candidates for case $n$
        \IF{$\mathrm{top1}(J_y)-\mathrm{top2}(J_y) \ge \delta$}
            \STATE $q^+ \leftarrow \arg\max_{q\in\mathcal{C}} J_y(q,c_n)$
            \FOR{each $q^-\in\mathcal{C}$ with $J_y(q^+,c_n)-J_y(q^-,c_n)\ge\delta$}
                \IF{$S_y(q^+,c_n) \ge S_y(q^-,c_n)-\varepsilon$}
                    \STATE $\mathcal{P} \leftarrow \mathcal{P}\cup\{(q^+,q^-)\}$
                \ENDIF
            \ENDFOR
        \ENDIF
    \ENDFOR
    \STATE train from $\theta$ for about one epoch on $\mathcal{P}$ using the full \hppo{} objective against $\theta_{\mathrm{ref}}$, saving round checkpoints $\Theta_r$
    \STATE $\theta_r \leftarrow \arg\max_{\vartheta\in\Theta_r}\operatorname{Validate}(\vartheta;\mathcal{V})$
    \STATE $J_r \leftarrow \operatorname{Validate}(\theta_r;\mathcal{V})$ using the same online metric and serving guidance scale
    \IF{$J_r > J^\star$}
        \STATE $J^\star \leftarrow J_r$;\quad $\theta \leftarrow \theta_r$;\quad $\theta^\star \leftarrow \theta_r$
    \ELSE
        \STATE break \quad (validation early stop)
    \ENDIF
\ENDFOR
\STATE Output: aligned policy $\theta^\star$
\end{algorithmic}
\end{algorithm}

For one domain, about 50{,}000 cases with eight candidates each yield 400{,}000 candidate queries, which a single machine labels with real retrieval in under an hour (about 127 queries per second, dominated by vector recall). A representative 100-step D1 preference update takes 1.3 minutes on eight GPUs. The iterated variants reuse the first-round labels for static candidates and relabel only the regenerated policy fan in later rounds. Appendix Tables~\ref{tab:training-cost} and~\ref{tab:inference-cost} report the measured downstream training and serving costs.

\subsection{Inference}

At inference the model conditions on a seed set of same-pattern item embeddings (in place of the training-time ordered prefix) and generates a query embedding. Generation starts from Gaussian noise projected onto the unit sphere and integrates the learned velocity field for exactly 30 Euler steps, renormalizing onto the sphere after each step and applying classifier-free guidance at scale $3.0$. The default 150-item condition prefix does not attend to the target, so the model prefills it once and reuses its key/value cache across the integration. Under the late-entry configuration ($\texttt{noise\_layer}=1$), each conditional and unconditional velocity evaluation traverses only the final of 28 transformer blocks; the other 27 blocks are not repeated at each step. The resulting query is $L_2$-normalized and searched in the same vector index as the baselines. Default serving uses one generated query and one approximate-nearest-neighbor request. Evaluation draws 20 independent noise samples per case to estimate query-level mean metrics; these samples are not combined into one serving request. Appendix Table~\ref{tab:inference-cost} reports the matched model and query-to-ANN latency. Unless otherwise stated, evaluation uses top-$K$ retrieval and computes all metrics on the retrieved items, not on predicted density scores.

\section{Experiments}

\subsection{Setup}

\subsubsection{Domains}

The D1--D4 experiments use anonymized multimodal content items sampled from the same proprietary corpus. A fixed upstream multimodal encoder integrates each item's visual content with its available associated textual signals. Before retrieval-model training, the encoder maps every item to the same frozen 1{,}536-dimensional representation. \mos{} therefore consumes sequences of frozen representations rather than raw content.

Within each domain, items satisfying a fixed binary attribute criterion form the target-positive pool. D1--D4 are evaluated as separate attribute-specific tasks rather than as a mutually exclusive four-class label space. The domains share the source corpus, representation space, ANN retrieval infrastructure, and evaluation protocol but use attribute-specific positive pools and separately fitted cluster spaces; equal numeric cluster identifiers across domains therefore have no shared semantics. Table~\ref{tab:domains} provides high-level attribute descriptions, Table~\ref{tab:d1-illustration} gives a concrete synthetic illustration of D1, and Table~\ref{tab:data} reports aggregate scale and split structure. Privacy constraints preclude releasing raw content, fine-grained collection metadata, and internal feature, scorer, or policy identifiers.

\begin{table}[!t]
\centering
\caption{Domain attribute families. Table~\ref{tab:d1-illustration} gives a concrete synthetic illustration.}
\label{tab:domains}
\begin{tabular}{@{}lp{0.28\linewidth}p{0.54\linewidth}@{}}
\hline
Domain & Attribute family & Attribute description \\
\hline
D1 & Engagement-inducing content. & An item is designed to induce clicks or user interactions. \\
D2 & Low-quality content. & An item is repetitive or provides little new information. \\
D3 & Non-original content. & Existing material is reposted with limited substantive changes. \\
D4 & Visually disturbing content. & An item contains disgusting or frightening visual elements. \\
\hline
\end{tabular}
\end{table}

\paragraph{Synthetic illustration.} Table~\ref{tab:d1-illustration} illustrates D1 concretely; every item in it is fabricated and does not reproduce a dataset item. D1 items are short videos of roughly $10$\,s to $5$\,min in both vertical and landscape formats. They usually carry no authored title, so the associated textual signal is on-screen text recovered by OCR rather than user-written metadata. A pattern is a fine-grained content and style region---one recurring generated-animation style, say---not a broad topic label.

\begin{table}[!t]
\centering
\setlength{\tabcolsep}{5pt}
\renewcommand{\arraystretch}{1.08}
\caption{Synthetic D1 illustration. Two fabricated patterns, each with one attribute-negative item and two attribute-positive items. No row reproduces a dataset item.}
\label{tab:d1-illustration}
\begin{tabular}{@{}>{\raggedright\arraybackslash}p{0.81\linewidth}>{\centering\arraybackslash}p{0.13\linewidth}@{}}
\hline
Synthetic item & Attribute \\
\hline
\multicolumn{2}{@{}>{\raggedright\arraybackslash}p{0.95\linewidth}@{}}{\textbf{P1. AI-generated animated serial.} \textit{Shared pattern:} vertical videos approximately 45--60 seconds long, with the same generated-animation style, recurring characters and color palette, episodic pacing, and burned-in narrative subtitles.} \\[2pt]
The episode plays entirely as generated animation. OCR-visible text contains only story dialogue and narration, with no request for viewer interaction. & negative \\
The same characters, animation style, and storyline are retained, but the video briefly cuts to live-action footage with an overlaid appeal for help that asks viewers to like and share before returning to the story. & positive \\
The same animation and subtitle style are retained, but an inserted segment makes a religious appeal accompanied by an OCR-visible request to comment in support before the story resumes. & positive \\
\hline
\multicolumn{2}{@{}>{\raggedright\arraybackslash}p{0.95\linewidth}@{}}{\textbf{P2. Tarot and astrology readings.} \textit{Shared pattern:} landscape videos approximately 60--90 seconds long, featuring a single presenter in a fixed front-facing setup, cards or astrology charts on the table, and consistent burned-in captions.} \\[2pt]
The presenter completes the reading and ends with the interpretation. OCR-visible captions summarize the reading without attaching any condition to the outcome. & negative \\
The same presenter, camera setup, and props are retained, but mid-reading the presenter states---and the captions repeat---that the outcome will take effect only for viewers who follow the account. & positive \\
The same format is retained, but the closing spoken line and caption promise that leaving a comment will bring luck to the viewer's family. & positive \\
\hline
\end{tabular}
\end{table}

Within P1 the items share the same generated-animation style, episodic form, and pacing; within P2 they share the same presentation format and props. In both fabricated patterns, the surrounding content and style define the Pattern, while an added engagement-soliciting device defines the Attribute---an inserted appeal-and-request segment in P1, and a conditional promise attached to a follow or comment in P2. The example therefore makes the task's crossed axes concrete: averaging same-Pattern seeds anchors the query to the shared content and style, but does not by itself favor items carrying the target Attribute. The matched D1 diagnostic in Appendix~\ref{app:three-trigger} quantitatively tests this motivating distinction.

\begin{table}[!t]
\centering
\caption{Strict-split data scale. Values are sequence rows, not distinct item IDs. Train rows exclude the held-out clusters; Eval-I and Eval-P are the item-holdout and pattern-holdout evaluation splits; Clusters is held-out over total pattern clusters.}
\label{tab:data}
\begin{tabular}{lrrrrr}
\hline
Domain & Train & Eval-I & Eval-P & Normal & Clusters \\
\hline
D1 & 1.54M & 156K & 7.8K & 10\% & 50/1000 \\
D2 & 5.95M & 163K & 32.8K & 20\% & 50/1000 \\
D3 & 9.22M & 163K & 48.7K & 20\% & 50/1000 \\
D4 & 478K & 36K & 1.8K & 20\% & 25/500 \\
\hline
\end{tabular}
\end{table}

\paragraph{Labeling and full-pool scoring.}

Online recall-density labels are used only to train and validate the density predictor. After validation, the predictor scores the full item pool and \mos{} construction uses the predicted scores to order items. The final retrieval metrics are never computed with predicted density: they are computed by querying the vector index and evaluating the actually retrieved top-$K$ items. This separation keeps labeling cost manageable while preserving an online-retrieval evaluation target.

\subsubsection{Baselines and Variants}

We compare the following models and query constructions, all evaluated under the strict item/cluster double-holdout protocol---the hardened configuration for the main results, and a lighter (fewer-case) configuration for the order ablation.
\begin{itemize}
    \item \textbf{Pretrained GR (Pre.\ GR)}: the shared pretrained continuous retriever.
    \item \textbf{Average pooling (Avg.)}: the mean of the input embeddings, used directly as the query.
    \item \textbf{ANCE-PRF (D1 only)}: a task-adapted baseline with a 12-layer, 87.54M-parameter FirstP query encoder trained on all 1.538M strict D1 cases using two complete ANN mining rounds, independent binary-triplet NLL, and a frozen item index. The inherited $k=3$ encoder consumes the first three unlabeled FirstP results and performs a second retrieval. Its matched $k=0$ control shares the complete initialization, 50k/2k PRF train/validation split, fixed negatives, and validation selection; all weight-updating stages exclude the formal holdouts.
    \item \textbf{\mos{} CPT/SFT}: shared multi-domain CPT followed by domain-specific tail-centroid SFT---the main pipeline. Its intermediate stage \emph{CPT only} (continuation pretraining without centroid SFT) is traced in the stage-wise progression (Table~\ref{tab:progress}).
    \item \textbf{Single-domain CPT+SFT}: the same pipeline with a per-domain CPT instead of the shared multi-domain CPT; used in the order ablation (Table~\ref{tab:order}).
    \item \textbf{\hppo{} and its ablations}: online-reward preference optimization initialized from the \mos{} SFT checkpoint, with static tail-centroid candidates and guidance-fan policy candidates labeled by online \shortJoint{}. The main \hppo{} row is the complete iterated Pareto-filtered procedure in every domain; \emph{off-policy} (single round) and \emph{adaptive} (iterated without the pair filter) isolate its mechanisms in Table~\ref{tab:hppo-abl}.
\end{itemize}
A further set of oracle constructions---sharp target queries from the true tail (\textbf{Target top-1/top-5} and \textbf{Target centroid})---exists only under an earlier non-strict, single-domain evaluation and is deferred to Appendix~\ref{app:original-results}; we use them only for their relative conclusions, as their absolute numbers are not comparable to the strict tables.

\subsubsection{Evaluation Protocol}

For each evaluation sequence, we generate a query embedding and retrieve top-$K$ nearest neighbors. We report raw attribute density, intersection density, conditional same-pattern purity among positives, and same-pattern share among all retrieved items. Space-constrained tables use the column names Attr, Same, Joint, and Cond for \metricIssue{}, \metricSame{}, \metricJoint{}, and \metricCond{} as defined in the problem setup. The AUC column reports within-pattern directional discrimination: the positives are the held-out high-metric tail items of a case and the negatives are input items of the same cluster, both scored by similarity to the generated queries. AUC therefore measures discrimination between high-metric tail items and input items within a pattern. In Table~\ref{tab:main}, average pooling ranges from $0.4565$ to $0.5631$, while \mos{} and \hppo{} exceed it in every domain--split cell. Reported values are query-level means; queries with zero retrieved attribute positives are skipped in the conditional mean, so aggregate columns do not multiply out exactly. For the strict-split experiments, every domain reserves both an item-holdout and a pattern-holdout (held-out-cluster) evaluation split, and all weight-updating stages exclude the held-out clusters. The main results (Tables~\ref{tab:main}, \ref{tab:progress}, \ref{tab:hppo-abl} and Figure~\ref{fig:sig}) use a hardened configuration: the within-pattern AUC over up to 1{,}000 cases per split, and the retrieval-density metrics over 500 cluster-stratified cases per split, with all models scored on the identical, deterministically sampled case set, so the rows are paired and directly comparable. The order ablation (Table~\ref{tab:order}) uses a lighter matched configuration with 50 stratified density cases per split; its values are comparable within the table but not in absolute magnitude to the hardened tables, and small density differences there should be interpreted cautiously given the smaller case sample. Significance for the main contrasts is assessed by paired bootstrap over the shared cases (Figure~\ref{fig:sig}).

Checkpoint selection for preference rows stays off the formal test sets. The iterated variants (adaptive and Pareto) select the round and checkpoint on a disjoint train-side validation slice labeled with true online \shortJoint{} and stop when a round does not improve it. The off-policy mechanism control instead uses one common fixed single-round training endpoint in every domain, with the same archived model evaluated on both formal holdouts. The checkpoint-drift trajectory (Figure~\ref{fig:drift}, Appendix~\ref{app:perdomain}) is retained as a diagnostic of unconstrained preference training, not as a formal-holdout selection rule.

\subsection{Main Results}

\subsubsection{Main Comparison}

On D1, ANCE-PRF obtains \shortJoint{} of $10.68\%/7.03\%$ on item/pattern holdout, compared with $10.81\%/7.14\%$ for its matched $k=0$ control and $17.36\%/13.84\%$ for \hppo{} (Table~\ref{tab:appdc} in Appendix~\ref{app:perdomain}). The paired ANCE-PRF-minus-\hppo{} gaps are $-6.68$~pp (95\% CI $[-7.78,-5.63]$) and $-6.82$~pp ($[-7.86,-5.80]$), respectively. Hence the additional two-pass baseline does not change the D1 ranking, and Top-3 pseudo-feedback does not improve its matched control under this protocol. The training trajectory suggests a task-specific explanation: a single binary relevance label conflates attribute-negative/same-pattern and attribute-positive/wrong-pattern items, which require different corrections. In contrast, \hppo{} forms query preferences from online \shortJoint{} and constrains pattern degradation through its Pareto filter. This pattern is consistent with better objective alignment in our dual-axis task.

Table~\ref{tab:main} reports the main result on strict item-holdout and pattern-holdout splits, with all models evaluated under one unified hardened protocol so the rows are directly comparable. Average pooling behaves exactly as the motivation predicts: it keeps the highest same-pattern share but stays in a low-attribute region, which caps its \shortJoint{}. \mos{} improves \shortJoint{} over both the pretrained retriever and average pooling on every domain and split. The complete Pareto-filtered \hppo{} procedure improves \shortJoint{} over its \mos{} initialization on seven of the eight domain--split cells, and all seven gains are paired-bootstrap significant. On D1, D2, and D3 the gains range from $+6.0$~pp (D1 item, 95\% CI $[+5.1,+6.9]$~pp) to over $+10$~pp (D3 item). The D4 item gain is smaller but significant ($+0.54$~pp, $[+0.22,+0.87]$~pp), while D4 pattern-holdout decreases by $0.35$~pp ($[-0.56,-0.15]$~pp); this is the observed boundary of the complete method. The detailed columns show that preference optimization raises raw attribute density while giving up part of same-pattern share and conditional purity. Table~\ref{tab:hppo-abl} therefore isolates the mechanisms directly: on D1--D3, Pareto filtering improves \shortJoint{} over the unfiltered adaptive policy on all six cells and raises \metricSame{} on four, while D4 receives no additional Joint benefit.

\begin{table}[!t]
\centering
\setlength{\tabcolsep}{4pt}
\caption{Main result on strict item-holdout and pattern-holdout splits. Avg.\ is average pooling of the input embeddings; Pre.\ GR is the shared pretrained retriever; \mos{} denotes shared multi-domain CPT plus domain-specific tail-centroid SFT; \hppo{} is the complete iterated Pareto-filtered procedure in every domain, with round and checkpoint selected only on disjoint train-side validation. All rows use one unified hardened evaluation protocol with 500 cluster-stratified cases per split shared across models, so values are directly comparable; \shortJoint{} differences are tested for significance by paired bootstrap (Figure~\ref{fig:sig}). Attr, Same, Joint, and Cond are percentages; AUC is a $0$--$1$ discrimination score.}
\label{tab:main}
\begin{tabular}{lllrrrrr}
\hline
Domain & Eval & Method & AUC & Attr & Same & Joint & Cond \\
\hline
D1 & Item & Avg. & 0.4793 & 12.63 & 90.51 & 8.29 & 87.65 \\
D1 & Item & Pre. GR & 0.4715 & 11.86 & 67.16 & 5.99 & 64.73 \\
D1 & Item & \mos{} & 0.5278 & 18.11 & 88.37 & 11.37 & 84.51 \\
D1 & Item & \hppo{} & 0.5992 & 35.34 & 76.29 & \textbf{17.36} & 65.90 \\
D1 & Pattern & Avg. & 0.4741 & 8.80 & 93.94 & 6.06 & 92.17 \\
D1 & Pattern & Pre. GR & 0.4659 & 8.58 & 73.30 & 4.58 & 68.70 \\
D1 & Pattern & \mos{} & 0.5171 & 11.55 & 89.92 & 7.74 & 83.92 \\
D1 & Pattern & \hppo{} & 0.5952 & 26.18 & 80.50 & \textbf{13.84} & 70.96 \\
\hline
D2 & Item & Avg. & 0.4977 & 3.89 & 78.75 & 2.81 & 73.82 \\
D2 & Item & Pre. GR & 0.4867 & 3.57 & 57.69 & 1.87 & 54.24 \\
D2 & Item & \mos{} & 0.5413 & 5.36 & 77.20 & 3.79 & 73.59 \\
D2 & Item & \hppo{} & 0.6068 & 22.47 & 52.68 & \textbf{10.28} & 46.00 \\
D2 & Pattern & Avg. & 0.5013 & 3.77 & 79.32 & 2.56 & 77.24 \\
D2 & Pattern & Pre. GR & 0.4886 & 3.66 & 58.11 & 1.98 & 60.70 \\
D2 & Pattern & \mos{} & 0.5426 & 7.16 & 75.94 & 4.99 & 72.49 \\
D2 & Pattern & \hppo{} & 0.6087 & 27.74 & 54.14 & \textbf{14.40} & 48.64 \\
\hline
D3 & Item & Avg. & 0.4565 & 8.97 & 93.17 & 8.03 & 92.56 \\
D3 & Item & Pre. GR & 0.4473 & 9.60 & 70.62 & 6.44 & 70.01 \\
D3 & Item & \mos{} & 0.5182 & 12.46 & 92.11 & 10.72 & 91.30 \\
D3 & Item & \hppo{} & 0.5916 & 42.08 & 66.97 & \textbf{21.42} & 59.85 \\
D3 & Pattern & Avg. & 0.4573 & 9.10 & 93.77 & 8.50 & 95.19 \\
D3 & Pattern & Pre. GR & 0.4525 & 9.14 & 73.56 & 6.75 & 75.86 \\
D3 & Pattern & \mos{} & 0.5054 & 11.77 & 91.06 & 10.72 & 92.74 \\
D3 & Pattern & \hppo{} & 0.5810 & 32.56 & 70.05 & \textbf{17.48} & 59.53 \\
\hline
D4 & Item & Avg. & 0.5631 & 16.21 & 87.53 & 9.53 & 82.93 \\
D4 & Item & Pre. GR & 0.5485 & 14.42 & 68.04 & 6.86 & 66.90 \\
D4 & Item & \mos{} & 0.6093 & 21.35 & 82.51 & 11.19 & 75.81 \\
D4 & Item & \hppo{} & 0.6356 & 26.28 & 75.47 & \textbf{11.73} & 66.16 \\
D4 & Pattern & Avg. & 0.5251 & 17.00 & 80.23 & 7.90 & 73.02 \\
D4 & Pattern & Pre. GR & 0.5117 & 14.69 & 66.44 & 5.82 & 62.13 \\
D4 & Pattern & \mos{} & 0.5871 & 22.67 & 73.66 & \textbf{8.99} & 65.16 \\
D4 & Pattern & \hppo{} & 0.6337 & 28.78 & 64.32 & 8.64 & 51.95 \\
\hline
\end{tabular}
\end{table}

\paragraph{Public reproducibility.} We repeat the task on the public Semi-Truths image dataset \citep{pal2024semitruths}, where both axes are externally inspectable. Dataset provenance defines the Attribute: an original image is positive and an AI-edited descendant is negative. A content cluster fitted on original images alone defines the Pattern, and every edited descendant inherits the Pattern of its source family; fitting the centers on originals prevents an edit from redefining the Pattern of its own item. Source families remain atomic across splits, and six complete Patterns are withheld from every weight-updating stage, including the label-free pretraining. Table~\ref{tab:public-main} reports the result.

The staged mechanism reproduces on the item holdout. Tail-centroid SFT moves the query off the seed centroid toward the attribute-dense region of the same Pattern, raising Attribute@$100$ from $46.93$ to $60.33\%$ at a cost of $2.2$ points of same-pattern share. \hppo{} replaces the proxy target with the true retrieved \shortJoint{} and raises Attribute@$100$ by a further $5.2$ points. Across the two stages the primary metric rises from $39.14$ to $50.30\%$.

On the six completely unseen Patterns, tail-centroid SFT raises Attribute@$100$ from $50.60$ to $67.86\%$, but same-pattern share falls from $61.42$ to $45.33\%$. Its \shortJoint{} gain over average pooling is $+0.58$~pp, and the six-Pattern-block 95\% interval $[-3.51,+3.58]$ includes zero (Appendix~\ref{app:public-semitruths}). We therefore restrict the tail-centroid stability claim to item-held-out cases from Patterns represented during training, and do not claim a statistically resolved pattern-preserving SFT improvement on completely unseen Patterns. Importantly, \hppo{} then adds a statistically resolved $+1.93$~pp in \shortJoint{} over SFT (95\% CI $[+1.20,+2.81]$) without further loss in the \metricSame{} point estimate ($45.33\to45.44\%$). Thus, even on this stringent split, the preference stage retains its intended role where the SFT step alone is not statistically resolved.

The public benchmark is intended to test effectiveness and transferability under externally inspectable definitions, not to serve as a scale- and diversity-matched replica of our production setting. Its image-only construction uses 128 Patterns, whereas the D1--D4 multimodal settings use 500--1,000 Patterns. Consistent with this lower effective Pattern diversity, average pooling already retrieves $46.93/50.60\%$ attribute-positive items on the public item/complete-Pattern holdouts, compared with $3.77$--$17.00\%$ on D1--D4; this stronger baseline leaves less headroom between methods. Nevertheless, \shortJoint{} retains the strict AvgPool $<$ SFT $<$ \hppo{} point ordering on both public holdouts, both adjacent gains are statistically resolved on the item holdout, and the \hppo{}-over-SFT gain remains resolved on the complete-Pattern holdout. Semi-Truths therefore provides public evidence of effectiveness and transferability, complementing D1--D4's evaluation of the larger and more heterogeneous operational regime.

\begin{table}[!t]
\centering
\caption{Public Semi-Truths benchmark, exact Top-$100$ retrieval. Attr, Same, and Joint are percentages. The Pattern split holds out six complete Patterns from every weight-updating stage. Appendix~\ref{app:public-semitruths} reports the full table, the Tail-Oracle reference, and Pattern-blocked confidence intervals.}
\label{tab:public-main}
\begin{tabular}{llrrr}
\hline
Eval & Method & Attr & Same & Joint \\
\hline
Item & Avg. & 46.93 & 80.68 & 39.14 \\
Item & SFT & 60.33 & 78.49 & 46.61 \\
Item & \hppo{} & 65.56 & 78.39 & \textbf{50.30} \\
\hline
Pattern & Avg. & 50.60 & 61.42 & 30.70 \\
Pattern & SFT & 67.86 & 45.33 & 31.28 \\
Pattern & \hppo{} & 71.95 & 45.44 & \textbf{33.21} \\
\hline
\end{tabular}
\end{table}

\subsubsection{Significance}

\begin{figure}[!t]
\centering
\includegraphics[width=0.93\textwidth]{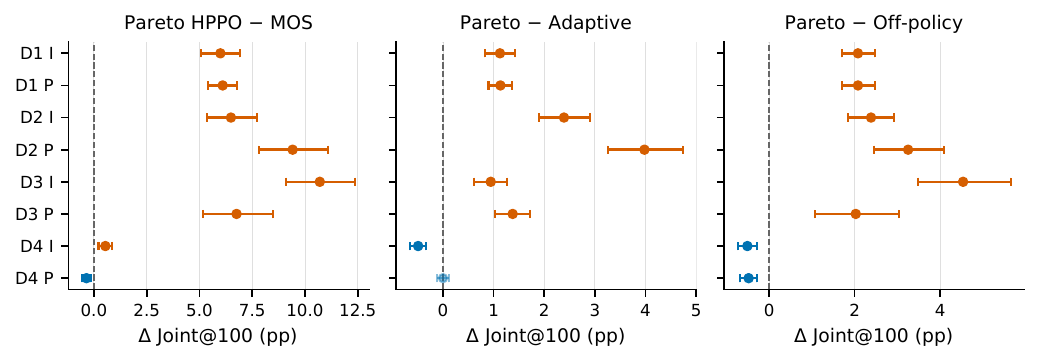}
\caption{Paired-bootstrap \shortJoint{} contrasts ($B{=}10{,}000$, $n{=}500$ shared cases). Each marker is the mean per-case difference with its $95\%$ confidence interval; the dashed line is zero. Rows are the eight domain--split cells (I${=}$item, P${=}$pattern). \emph{Left}: the complete Pareto-filtered \hppo{} procedure minus its \mos{} initialization. \emph{Middle} and \emph{right}: Pareto versus the iteration-only Adaptive variant and the common-fixed-endpoint Off-policy variant. Orange favors the first method in each title, blue the second; the one faded marker (Pareto vs.\ Adaptive on D4 pattern) is the only interval that includes zero.}
\label{fig:sig}
\end{figure}

Because every model is evaluated on the identical set of cluster-stratified cases (deterministic sampling, verified $100\%$ case-id overlap) and the per-query intersection values are stored, the comparisons are \emph{paired}. We resample the shared cases with replacement ($B=10{,}000$) and report the bootstrap distribution of each per-case mean difference. Figure~\ref{fig:sig} shows the three key \shortJoint{} contrasts. The complete Pareto-filtered \hppo{} procedure is significantly above \mos{} on seven cells and significantly below it on D4 pattern-holdout. Pareto is also significantly above both the iteration-only policy (\emph{adaptive}) and the fixed-$100$ single-round \emph{off-policy} control on all six D1/D2/D3 cells. On D4 it trails off-policy on both splits, trails adaptive on item holdout, and has no statistically resolved difference from adaptive on pattern holdout. Paired testing resolves close contrasts: the marginal confidence intervals of the Pareto and adaptive runs overlap (e.g.\ D1 item \shortJoint{} $17.4\,[15.8,18.9]$ vs.\ $16.2\,[14.7,17.7]$, in \%), yet their paired difference $+1.1\,[+0.8,+1.4]$~pp excludes zero because the paired comparison controls for shared case difficulty. These bootstrap intervals quantify evaluation-sampling uncertainty. The \shortJoint{} improvement is also stable across independent training runs: re-training the iterated-Pareto preference stage with three random seeds---each re-samples the training cases, regenerates the candidates, and re-optimizes, while the evaluation set is held fixed---keeps every D1--D3 cell well above \mos{} on both splits (Table~\ref{tab:seed}).

\begin{table}[!t]\centering\setlength{\tabcolsep}{4pt}
\caption{Seed robustness of the iterated-Pareto \hppo{} \shortJoint{} (\%, higher is better) across three independent training runs with different random seeds (each re-samples training cases, regenerates candidates, and re-optimizes; the evaluation set is held fixed), with \mos{} for reference. Seed~1 is the run reported in Table~\ref{tab:main}. The improvement over \mos{} is stable across seeds on D1--D3; D4 remains variable and is analyzed in Appendix~\ref{app:dbexception}.}
\label{tab:seed}
\begin{tabular}{llrrrrr}
\hline
Domain & Split & \mos{} & Seed 1 & Seed 2 & Seed 3 & Mean \\
\hline
D1 & Item & 11.37 & 17.36 & 17.99 & 16.66 & 17.34 \\
D1 & Pattern & 7.74 & 13.84 & 15.94 & 14.43 & 14.74 \\
D2 & Item & 3.79 & 10.28 & 10.17 & 8.62 & 9.69 \\
D2 & Pattern & 4.99 & 14.40 & 13.70 & 11.32 & 13.14 \\
D3 & Item & 10.72 & 21.42 & 20.77 & 20.91 & 21.03 \\
D3 & Pattern & 10.72 & 17.48 & 15.88 & 15.79 & 16.38 \\
D4 & Item & 11.19 & 11.73 & 11.94 & 11.42 & 11.70 \\
D4 & Pattern & 8.99 & 8.64 & 9.96 & 9.58 & 9.39 \\
\hline
\end{tabular}
\end{table}

\subsubsection{Stage-wise Progress}

Table~\ref{tab:progress} traces the primary \shortJoint{} metric across the four training stages on the strict splits, all under the unified hardened protocol. The base initialization retrieves essentially no same-pattern positives; large-scale raw-sequence pretraining establishes a usable retrieval prior; the shared \mos{} CPT improves seven of the eight point estimates over pretraining, with D4 pattern-holdout decreasing from $5.82$ to $5.32\%$; domain-specific tail-centroid SFT provides the main supervised jump; and the complete Pareto-filtered \hppo{} procedure further improves seven of the eight cells (see Appendix~\ref{app:dbexception} for the D4 boundary).

\begin{table}[!t]
\centering
\setlength{\tabcolsep}{3pt}
\caption{Stage-wise progress on the primary \shortJoint{} metric. Base is the base initialization, Pre.\ is raw-sequence pretraining, CPT is the shared multi-domain \mos{} CPT, SFT is domain-specific tail-centroid SFT from the shared CPT, and \hppo{} is the complete iterated Pareto-filtered procedure in every domain. All stages are evaluated under the unified hardened protocol of Table~\ref{tab:main}. Values are \shortJoint{} in percent.}
\label{tab:progress}
\begin{tabular}{llrrrrr}
\hline
Eval & Dom. & Base & Pre. & CPT & SFT & \hppo{} \\
\hline
Item & D1 & 0.00 & 5.99 & 7.52 & 11.37 & 17.36 \\
Item & D2 & 0.00 & 1.87 & 2.26 & 3.79 & 10.28 \\
Item & D3 & 0.02 & 6.44 & 7.47 & 10.72 & 21.42 \\
Item & D4 & 0.00 & 6.86 & 7.30 & 11.19 & 11.73 \\
\hline
Pattern & D1 & 0.00 & 4.58 & 5.03 & 7.74 & 13.84 \\
Pattern & D2 & 0.00 & 1.98 & 3.09 & 4.99 & 14.40 \\
Pattern & D3 & 0.01 & 6.75 & 7.61 & 10.72 & 17.48 \\
Pattern & D4 & 0.00 & 5.82 & 5.32 & 8.99 & 8.64 \\
\hline
\end{tabular}
\end{table}

\subsection{Ablations and Analysis}

\subsubsection{Metric Ordering}

The central hypothesis is that ordering same-pattern sequences by the proxy metric teaches a directional transformation. We therefore use controlled order ablations with identical data scale, split, model initialization, and training configuration: ascending density, random order, and descending density. These ablations use the single-domain CPT+SFT instantiation so that sequence order is the only varying factor. Table~\ref{tab:order} reports the primary \shortJoint{} metric on both item-holdout and pattern-holdout evaluation sets. Ascending density is the best row in every domain and split, and also exceeds the base initialization, the shared pretrained checkpoint, and continuation pretraining without SFT.

\begin{table}[!t]
\centering
\setlength{\tabcolsep}{4pt}
\caption{Matched order ablation on the primary \shortJoint{} metric (in percent), in the single-domain instantiation so that sequence order is the only varying factor. Item and Pattern denote item-holdout and pattern-holdout evaluation sets. Base is the base initialization, Pre.\ is the shared pretrained GR checkpoint, and CPT is continuation pretraining without centroid SFT. Because this table uses the single-domain instantiation under the light evaluation protocol, its absolute values are comparable only within the table (Asc.\ vs.\ Rand.\ vs.\ Desc.) and not across to Table~\ref{tab:main} or Table~\ref{tab:progress}, which report the shared multi-domain pipeline under the hardened protocol.}
\label{tab:order}
\begin{tabular}{llrrrrrr}
\hline
Eval & Dom. & Base & Pre. & CPT & Desc. & Rand. & Asc. \\
\hline
Item & D1 & 0.00 & 5.40 & 6.28 & 7.37 & 7.87 & \textbf{9.25} \\
Item & D2 & 0.00 & 2.71 & 3.03 & 3.89 & 3.52 & \textbf{4.11} \\
Item & D3 & 0.00 & 6.07 & 7.36 & 7.22 & 7.19 & \textbf{11.06} \\
Item & D4 & 0.00 & 6.73 & 6.91 & 7.88 & 8.98 & \textbf{11.40} \\
\hline
Pattern & D1 & 0.01 & 4.39 & 4.81 & 4.94 & 5.96 & \textbf{7.07} \\
Pattern & D2 & 0.00 & 1.54 & 2.16 & 1.74 & 1.46 & \textbf{2.40} \\
Pattern & D3 & 0.01 & 6.79 & 6.91 & 8.25 & 8.53 & \textbf{9.80} \\
Pattern & D4 & 0.00 & 6.14 & 5.98 & 5.42 & 5.70 & \textbf{9.32} \\
\hline
\end{tabular}
\end{table}

\subsubsection{HPPO Ablations}

\paragraph{Preference variants.} Table~\ref{tab:hppo-abl} treats off-policy and Adaptive as mechanism ablations of the complete Pareto-filtered \hppo{} procedure. The Off-policy control fixes the common $100$-step endpoint in every domain and evaluates the same model on both formal holdouts; it improves the \shortJoint{} point estimate over SFT on all eight cells. Regenerating candidates from the current policy (Adaptive) helps D1 and D3 but is neutral or negative on D2 and D4, so iteration is not uniformly beneficial by itself. Adding the Pareto pair filter changes this pattern: on D1--D3 it raises \shortJoint{} over Adaptive on all six cells and raises \metricSame{} on four (both D2 and D3 splits). Relative to the fixed Off-policy control, Pareto also raises \shortJoint{} on all six D1--D3 cells, although its final-policy \metricSame{} is higher only on D2; the pair-level constraint therefore narrows the training search rather than guaranteeing a global Same increase. D4 is the boundary: strict Pareto trails fixed Off-policy by $0.51$~pp on item holdout and $0.48$~pp on pattern holdout (paired $95\%$ CIs $[-0.73,-0.29]$ and $[-0.68,-0.29]$~pp) and has lower \metricSame{} on both. Appendix~\ref{app:dbexception} analyzes the data scale, margin-qualified preference signal, candidate headroom, and seed variability behind this failure case.

\paragraph{Candidate complementarity.} A second ablation on D1 toggles the candidate families inside the off-policy regime, with all rows sharing the same cases, online labels, and final $400$-step training budget. The full hybrid set (static tail centroids plus policy samples) reaches \shortJoint{} $15.28\%$ item / $12.05\%$ pattern, above static-only ($14.68$/$11.05$) and policy-only ($13.40$/$10.08$); Table~\ref{tab:cand} lists the precise values. Both families contribute, so the hybrid design is justified. This ablation also motivated the guidance fan: after point-target tail-centroid SFT the conditional generator is nearly deterministic, so independently sampled policy queries collapse to one effective candidate; fanning the guidance scale of the current policy restores the diversity that makes iteration and the pair filter effective.

\begin{table}[!t]
\centering
\setlength{\tabcolsep}{3pt}
\caption{\hppo{} mechanism ablation on the primary \shortJoint{} metric. SFT is the \mos{} initialization. Off-pol.\ collects candidates once from the frozen SFT policy plus static tail centroids and uses the common fixed $100$-step endpoint, with the same model evaluated on both holdouts. Hyb-adapt.\ and Hyb-pareto.\ regenerate guidance-fan candidates from the current policy each round and use disjoint train-side validation, respectively without and with the Pareto pair filter. Best per row in bold. Values are \shortJoint{} in percent.}
\label{tab:hppo-abl}
\begin{tabular}{llrrrr}
\hline
Eval & Dom. & SFT & Off-pol. & Hyb-adapt. & Hyb-pareto \\
\hline
Item & D1 & 11.37 & 15.28 & 16.24 & \textbf{17.36} \\
Item & D2 & 3.79 & 7.90 & 7.89 & \textbf{10.28} \\
Item & D3 & 10.72 & 16.88 & 20.48 & \textbf{21.42} \\
Item & D4 & 11.19 & \textbf{12.24} & 12.22 & 11.73 \\
\hline
Pat. & D1 & 7.74 & 11.77 & 12.71 & \textbf{13.84} \\
Pat. & D2 & 4.99 & 11.15 & 10.42 & \textbf{14.40} \\
Pat. & D3 & 10.72 & 15.45 & 16.10 & \textbf{17.48} \\
Pat. & D4 & 8.99 & \textbf{9.12} & 8.64 & 8.64 \\
\hline
\end{tabular}
\end{table}

\paragraph{On-policy iteration at fixed compute.} The off-policy-vs-adaptive comparison above is not step-matched. We therefore use a separate controlled D1 run to isolate candidate regeneration from the amount of training: the total preference budget is fixed at 100 steps and only the number of on-policy rounds varies. Table~\ref{tab:rollout} shows \shortJoint{} rising monotonically with the number of rounds on both splits at identical data and identical total steps. This diagnostic is interpreted only through its within-table contrast; its one-round reference is not the archived Off-policy checkpoint used in Table~\ref{tab:hppo-abl}.

\begin{table}[!t]
\centering
\setlength{\tabcolsep}{6pt}
\caption{Candidate-pool complementarity (D1, off-policy regime) at the common final $400$-step training budget. Static = four tail centroids; policy = four guidance-fan policy samples; hybrid = both families. Values are \shortJoint{} in percent.}
\label{tab:cand}
\begin{tabular}{lrr}
\hline
Candidate pool & Item & Pattern \\
\hline
Static-only (4 tail centroids) & 14.68 & 11.05 \\
Policy-only (4 fan samples) & 13.40 & 10.08 \\
Hybrid (static $+$ policy) & \textbf{15.28} & \textbf{12.05} \\
\hline
\end{tabular}
\end{table}

\begin{table}[!t]
\centering
\setlength{\tabcolsep}{6pt}
\caption{Separate on-policy-iteration diagnostic at fixed total compute (D1, hardened): same data and the same 100 total preference-training steps, varying only the number of candidate-regeneration rounds. Values are comparable within this table and are not substituted into the mechanism controls in Table~\ref{tab:hppo-abl}. \shortJoint{} is in percent.}
\label{tab:rollout}
\begin{tabular}{lrr}
\hline
Schedule (rounds$\times$steps) & Item & Pattern \\
\hline
$1\times100$ (single round) & 12.96 & 9.14 \\
$2\times50$ & 13.37 & 9.57 \\
$4\times25$ & \textbf{13.74} & \textbf{10.02} \\
\hline
\end{tabular}
\end{table}

\paragraph{Flow-surrogate reliability.} We directly test the score used by \hppo{} on 1{,}000 held-out D1 cases per split, using the cached eight-candidate hybrid pools and their true online rewards. For every fixed candidate, we repeat the policy and reference score 64 times. Across all retained pairs, reference-relative accuracy is $82.70\%\,[80.69,84.64]$ on item holdout and $73.82\%\,[71.28,76.36]$ on pattern holdout (chance: $50\%$). Table~\ref{tab:surrogate-main} shows when the surrogate is reliable: as the true reward gap increases, accuracy rises monotonically and both one-draw and eight-draw sign flips decrease. The static candidates, which are not sampled from the generative policy, also show no variance inflation: their mean score standard deviation is $0.76\times$ (item) and $0.81\times$ (pattern) that of policy-generated candidates. These results validate the property required by \hppo{}---reliable reference-relative preference ordering---without interpreting the score as an exact likelihood. Appendix~\ref{app:bon-analyses} gives the repeated-score protocol and candidate-level stability results.

\begin{table}[!t]
\centering
\setlength{\tabcolsep}{3pt}
\caption{Direct validation of the reference-relative flow-matching surrogate on D1 held-out candidates. Reward-separated pairs pass the training margin and strict Pareto filter. Acc. is agreement with the true online-reward ordering using the 64-draw mean (chance: $50\%$; case-bootstrap $95\%$ CI). Flip@1/8 compare one/eight-draw estimates with an independent 32-draw consensus (lower is better). Repeated draws are diagnostic only; \hppo{} training uses one draw per forward pass.}
\label{tab:surrogate-main}
\begin{tabular}{llrrccc}
\hline
Eval & Reward gap & Cases & Pairs & Acc. (95\% CI) & Flip@1 & Flip@8 \\
\hline
Item & $[0.03,0.05)$ & 422 & 816 & $72.06\,[68.11,75.87]$ & 28.24 & 22.21 \\
Item & $[0.05,0.10)$ & 436 & 1{,}108 & $82.22\,[79.19,85.08]$ & 21.95 & 15.30 \\
Item & $\geq0.10$ & 333 & 1{,}198 & $90.40\,[87.62,92.84]$ & 14.28 & 7.43 \\
\hline
Pattern & $[0.03,0.05)$ & 371 & 727 & $64.65\,[59.56,69.54]$ & 24.78 & 18.54 \\
Pattern & $[0.05,0.10)$ & 401 & 859 & $69.97\,[65.46,74.29]$ & 24.39 & 17.49 \\
Pattern & $\geq0.10$ & 300 & 1{,}179 & $82.27\,[78.76,85.52]$ & 18.00 & 11.41 \\
\hline
\end{tabular}
\end{table}

\subsubsection{Best-of-N Oracle: Generation vs.\ Selection}

A natural question is whether \hppo{} merely learns to select a good candidate from the menu it already had at initialization, or whether it learns to \emph{generate} better queries. We answer it with a best-of-$N$ oracle analysis on 1{,}000 held-out cases per split. For a fixed policy we form the same eight candidates used in training (four static tail centroids and four guidance-fan policy samples), label each with true online \shortJoint{}, and compare the policy's single served query against per-case oracles that pick the best candidate in hindsight. Table~\ref{tab:bon} reports the result. On D1, D2, and D3 the \hppo{} policy's \emph{single} generated query already exceeds the SFT policy's full eight-candidate oracle: \hppo{} has moved the generator itself, not just improved selection over the initialization's menu. The effect is starkest on D2, where the static centroids are nearly useless ($\shortJoint{}\le2\%$) yet the \hppo{} single query reaches $5.6\%$ (item) and $8.0\%$ (pattern). An eight-candidate oracle over the \hppo{} policy adds a further $15$--$27\%$ relative, indicating real headroom for a future inference-time reranker. On D4 the served query beats the SFT oracle only on the item split, and its generation--selection headroom is small on both splits---reflecting the limited D4 reward signal.

\begin{table}[!t]
\centering
\setlength{\tabcolsep}{3pt}
\caption{Best-of-$N$ oracle on held-out cases (\shortJoint{}, $1{,}000$ cases/split). SFT-oracle and \hppo{}-oracle pick the best of eight candidates per case in hindsight; \hppo{}-single is the policy's served query. On D1--D3 and on the D4 item split, the \hppo{} single query exceeds the SFT eight-candidate oracle. Values are \shortJoint{} in percent.}
\label{tab:bon}
\begin{tabular}{llrrr}
\hline
Eval & Dom. & SFT-oracle & \hppo{}-single & \hppo{}-oracle \\
\hline
Item & D1 & 10.20 & \textbf{12.60} & 15.70 \\
Item & D2 & 1.90 & \textbf{5.60} & 6.80 \\
Item & D3 & 9.70 & \textbf{15.70} & 19.10 \\
Item & D4 & 11.90 & 12.60 & 13.70 \\
\hline
Pat. & D1 & 10.50 & \textbf{13.50} & 17.10 \\
Pat. & D2 & 1.70 & \textbf{8.00} & 9.20 \\
Pat. & D3 & 8.00 & \textbf{12.80} & 15.90 \\
Pat. & D4 & 8.20 & 7.20 & 8.60 \\
\hline
\end{tabular}
\end{table}

\section{Conclusion}

We presented \method{}, a staged training framework for attribute-seeking and pattern-preserving generative retrieval. Starting from large-scale raw-sequence pretraining, \mos{} converts sparse online metric labels into metric-ordered in-pattern trajectories, trains one shared multi-domain continuation-pretrained retriever and a lightweight domain-specific tail-centroid SFT, and \hppo{} aligns the final query distribution with the online intersection metric through iterated reference-anchored preference optimization over hybrid candidates, using a tradeoff-aware Pareto pair filter to constrain purity-sacrificing training pairs. Across four large-scale domains, the \mos{} training stages improve the primary intersection metric over a strong pretrained retriever, and the complete Pareto-filtered \hppo{} procedure adds paired-bootstrap-significant gains on seven of the eight domain--split cells. Controlled order ablations, a mechanism ablation, and the candidate-policy ablation explain where the gains come from: Pareto filtering raises \shortJoint{} over the unfiltered Adaptive policy on every D1--D3 cell and raises \metricSame{} on four.

\paragraph{Limitations.}
\method{} requires a frozen item-embedding space and a production attribute scorer, and it derives the fine-grained pattern by clustering those embeddings---a scalable but approximate, rather than human-labeled, notion of a pattern. The quantitative validation in Appendix~\ref{app:patternproxy} shows strong non-random alignment with independently produced human annotations; because the proxy clusters are finer-grained than the human patterns, this evidence supports the proxy interpretation rather than one-to-one taxonomy recovery. The metric predictor is used only to order the training sequences and not for final evaluation, so its errors affect data construction rather than the reported metrics. The method assumes that useful in-pattern attribute improvement is available. Geometric measurements show stronger alignment between the label direction and dominant style axis on D4 (median $\cos^2 = 0.54$) than on D1--D3 (median $\cos^2 \le 0.10$). This association is consistent with the more difficult attribute--pattern tradeoff observed on D4. The Pareto filter constrains retrieved same-pattern share at pair construction, rather than geometric style displacement itself; Appendix~\ref{app:dbexception} reports the diagnostic evidence. The frozen-embedding requirement enables efficient large-scale retrieval with precomputed indexes but precludes end-to-end joint optimization of embeddings and retrieval policy; such joint training is a natural direction for future work when encoder retraining is feasible.

\paragraph{Broader impact.}
By finding more same-pattern examples of an attribute, the method can improve quality, safety, and integrity workflows, but the same capability could amplify biased or incorrectly defined attribute scorers; deployment should pair it with scorer auditing, human review for sensitive domains, and monitoring for pattern drift or over-retrieval of narrow content groups.

\bibliographystyle{tmlr}
\bibliography{main}

\clearpage
\appendix

This appendix collects architecture and implementation details (A), training and evaluation configuration (B), full per-domain results with confidence intervals (C), the matched three-trigger motivation diagnostic (D), best-of-$N$, candidate, flow-surrogate, and guidance analyses (E), pattern-proxy and metric-predictor validation (F), the earlier full-protocol single-domain results (G), the public Semi-Truths benchmark (H), negative and diagnostic results (I), and reproducibility and anonymization notes (J).

\section{Architecture and Implementation Details}

\paragraph{Unified diffusion transformer.} The generator is a single diffusion-transformer backbone trained with a flow-matching objective. The condition items and the noised target token(s) form one sequence that is processed by one transformer; there is no separate condition encoder feeding a pooled vector to a generation head. Each item contributes one token (its embedding projected to the hidden width plus a positional encoding and a learned role marker that distinguishes condition from target positions); a binary condition mask flags valid positions.

\paragraph{Structured attention mask.} The attention pattern has four blocks. (i) Condition--condition is causal (lower triangular), so the conditioning prefix is autoregressive. (ii) Condition--target is empty: condition positions never attend to targets, so the condition can be encoded once and cached. (iii) Target--condition is prefix-visible: each target attends to the condition prefix up to its own position. (iv) Target--target is diagonal: a target attends only to itself, so several targets denoise/generate independently in a single pass (used by the dense-prefix CPT objective; SFT uses a single target). Because each target attends to the full condition sequence rather than a pooled summary, fine-grained information from the seed items is preserved; this is the design difference from a lightweight single-vector-conditioned generation head.

\paragraph{Flow matching, normalization, and guidance.} The network predicts a flow-matching velocity on the $L_2$-normalized $d$-dimensional embedding sphere, using the spherical (Riemannian) flow-matching instantiation \citep{chen2024riemannian} with geodesic paths and the Jacobi weight $w(t,\omega)=[\sin(t\omega)/(t\omega)]^2$ defined in Equation~\eqref{eq:fm}. Noise and targets, including tail means, are normalized to unit length before path construction. The policy, frozen reference, and tail-anchor terms use this weighted loss; only the preference scores are divided by $d$. The diffusion timestep is injected at target positions through adaptive layer normalization (AdaLN). Classifier-free guidance is implemented as a condition/attention dropout during training, and the serving guidance scale is $3.0$ (validated near-optimal in Appendix~\ref{app:bon-analyses}). Inference---spherical ODE integration with key/value caching of the condition prefix, then approximate-nearest-neighbor search with no quantization or identifier decoding---is described in the main-text Inference subsection.

\paragraph{Model scale and measured inference cost.} The pretrained, \mos{}, and \hppo{} checkpoints use the same DiT architecture; the later stages change its weights but not its size. The model has 621.3M parameters (width 1024, 28 blocks, 16 attention heads, 8 key/value heads, and FFN width 3072), and its bf16 safetensors checkpoint occupies 1.24 GB (1.16 GiB). Table~\ref{tab:inference-cost} measures the formal one-query serving path on one NVIDIA H100 80GB GPU with 32 real prefixes from distinct clusters in the strict pattern-holdout split, each truncated to 150 items, CFG 3.0, and 30 Euler steps. We use PyTorch eager execution, place condition embeddings on the GPU before timing, run 20 warmups, and collect three timing blocks (100 requests each for DiT and 500 each for Avg). The ANN and query-to-ANN columns use 100 alternating requests per method to the same top-100 backend. Checkpoint loading and condition-embedding fetch are excluded from request latency; both methods share the latter. FLOPs are PyTorch operator counts for matrix multiplications, linear layers, and attention on the same batch-one execution graph; elementwise operations and normalization are excluded.

\begin{table}[!h]
\centering
\caption{Matched inference cost. Latencies are P50/P95 milliseconds. Query build measures only query construction; Shared ANN pools 200 alternating requests because both methods use the same backend; Query+ANN measures from resident condition embeddings through one top-100 ANN response. Peak allocation includes resident weights and request tensors; Avg is measured without loading DiT.}
\label{tab:inference-cost}
\begin{tabular}{lrr}
\hline
Metric & Avg. & \mos{} / \hppo{} \\
\hline
Params & 0 & 621.3M \\
FLOPs/query & $<0.001$G & 141.79G \\
Query build & 0.09 / 0.10 & 104.04 / 105.46 \\
Shared ANN & 349.90 / 387.97 & 349.90 / 387.97 \\
Query+ANN & 350.06 / 412.25 & 456.64 / 495.86 \\
Peak GiB & 0.002 & 1.210 \\
\hline
\end{tabular}
\end{table}

At batch size 32, DiT query construction reaches 278.0 queries/s; average pooling remains negligible relative to the shared ANN. Thus DiT adds a paired median 107.35 ms in this configuration, increasing the median query-to-ANN response from 350.06 to 456.64 ms (30.4\%). A text DPR pipeline would require replacing and retraining the embedding/index stack, while HyDE additionally generates and re-encodes text. We therefore use Avg as the matched non-generative baseline, with the same multimodal seed embeddings, frozen item space, index, and ANN request.

\section{Training and Evaluation Configuration}

\paragraph{Default constants.} Unless stated otherwise, the retrieval depth is $K=100$, the embedding dimension $d=1536$, the metric-ordered sequence length $L=200$, and the input prefix $m=150$ (so the high-metric tail has $L-m=50$ items). Here $K$ is the number of ANN results used to compute retrieval metrics. It is distinct from $\mathcal{K}$ in the \hppo{} candidate construction below: those values count high-metric tail items averaged into a static centroid, not retrieved neighbors.

\paragraph{Stages.} The downstream MORPH pipeline initializes from a pretrained continuous retriever obtained by large-scale raw-sequence pretraining over naturally ordered item sequences. The shared multi-domain \mos{} CPT mixes the four domains with repetition weights D1$\times2$, D2$\times1$, D3$\times1$, D4$\times3$, one epoch over the pool: global batch $4096$, per-device batch $16$, gradient accumulation $4$, learning rate $5\times10^{-4}$, warmup ratio $0.01$, minimum learning-rate ratio $0.1$ ($4806$ steps). Domain-specific tail-centroid SFT initializes from the shared CPT: global batch $4096$, per-device batch $64$, gradient accumulation $1$, learning rate $5\times10^{-5}$, warmup ratio $0.01$, tail-centroid target length $L-m$; D1/D2/D3 one epoch ($377/1452/2250$ steps), D4 two epochs ($234$ steps).

\paragraph{Measured downstream training cost.} Table~\ref{tab:training-cost} reports wall time from the formal strict-run logs. The shared CPT is amortized across the four domains, and each SFT row is a domain-specific continuation from that checkpoint. The D1 \hppo{} row measures one representative 100-step preference update; online candidate labeling is the separate recall-dominated process reported in the main text. The raw pretrained initialization was reused rather than rerun, so the table reports the controlled downstream stages.

\begin{table}[!h]
\centering
\caption{Measured downstream training cost for the formal strict pipeline. GPU-hours equal wall time multiplied by the allocated GPU count.}
\label{tab:training-cost}
\begin{tabular}{lrrrr}
\hline
Stage & Steps & GPUs & Wall time (h) & GPU-hours \\
\hline
Shared multi-domain CPT & 4,806 & 64 & 5.07 & 324.6 \\
D1 tail-centroid SFT & 377 & 64 & 0.50 & 32.0 \\
D2 tail-centroid SFT & 1,452 & 64 & 1.51 & 96.4 \\
D3 tail-centroid SFT & 2,250 & 64 & 2.31 & 147.7 \\
D4 tail-centroid SFT & 234 & 64 & 1.43 & 91.8 \\
D1 \hppo{} preference update & 100 & 8 & 0.022 & 0.17 \\
\hline
\end{tabular}
\end{table}

\paragraph{HPPO.} Candidates per case are four static tail centroids (top-$k$ averages over $\mathcal{K}=\{10,25,35,50\}$) and four policy samples from a guidance-scale fan over $\mathcal{S}=\{1,2,3,6\}$. The reward is true online \shortJoint{}. The loss is pairwise reference-anchored (DPO-style) with the frozen SFT model as reference, margin $\delta=0.03$ on the reward gap, and a tail-centroid anchor of weight $0.1$; learning rate $1\times10^{-5}$. The iterated variants run short rounds (about one epoch of the kept pairs, roughly $100$ steps), regenerate the policy fan from the current policy each round, select the round checkpoint on a disjoint train-side validation slice (held-out tail cases generated at the serving scale and labeled with true online \shortJoint{}), and accept a round only if its best saved checkpoint strictly improves the best retained score, initially the SFT validation score. The final D1--D4 Pareto runs used this strict rule without grace-period acceptance of non-improving rounds. The Off-policy control in Table~\ref{tab:hppo-abl} uses the archived $100$-step endpoint for every domain and evaluates that same checkpoint on both formal holdouts. The candidate-family ablation in Table~\ref{tab:cand} is a separate fixed-budget comparison at $400$ steps.

\paragraph{ANCE-PRF baseline.} The D1 FirstP query encoder uses 12 Transformer layers (hidden width 768; 87.54M trainable parameters), the fixed 1536-dimensional item embeddings, and LAMB at learning rate $10^{-6}$. We train on 1,538,090 strict cases with a 64-case global batch and two complete ANN hard-negative refreshes. Each case retains four positive targets and 16 deterministically shuffled fixed negatives; every positive--negative pair contributes an independent two-logit NLL. Joint@100 on a disjoint 5,000-case train-side validation set selects the first HN checkpoint at 20\% of an epoch; the second ANN refresh does not improve it. Matched $k=0$ and $k=3$ branches inherit the complete selected FirstP, train on exactly 50,000 cases with LAMB at $10^{-5}$, and select on a disjoint 2,000-case validation set. Both run for at least two epochs, and both select the common initialization because every later checkpoint is worse. The mining trajectory exposes the dual-axis supervision issue: 74.24\% of HN1 negatives are attribute-negative/same-pattern, and continued NLL training raises Attribute while sharply reducing Same; after HN2 changes this share to 46.76\% and introduces 26.96\% attribute-positive/wrong-pattern negatives, the early update reverses that pressure but still does not improve Joint. Both groups are Joint-negative yet call for different corrections, making the resulting update sensitive to the mined mixture. In contrast, \hppo{} compares complete query candidates using online \shortJoint{} and retains Pareto pairs that control pattern degradation. Formal item/pattern evaluation starts only after freezing these checkpoints and uses the same hardened cases as the other D1 rows.

\paragraph{Evaluation protocols.} The hardened protocol (main results) uses up to $1{,}000$ AUC cases and $500$ cluster-stratified density cases per split, with all models scored on one deterministic shared case sample (paired). The light protocol (order/CPT ablations) uses $50$ density cases per split; small differences require caution because of the smaller case sample. Significance is paired bootstrap over the shared cases, $B=10{,}000$ resamples of the per-case differences.

\paragraph{Feasibility.} For one domain, $50{,}000$ cases $\times$ $8$ candidates $=400{,}000$ candidate queries are labeled by real retrieval in under an hour on one machine ($\approx127$ queries/s, recall-dominated). Margin filtering keeps $\approx12\%$ of cases; a preference round trains in minutes on eight GPUs; a full iterated run is a few rounds.

\section{Full Per-Domain Results with Confidence Intervals}
\label{app:perdomain}

Tables~\ref{tab:appdc}, \ref{tab:appdp}, \ref{tab:appdu}, and \ref{tab:appdb} give every variant on every split with all five metrics; the \shortJoint{} cell carries its $95\%$ paired-bootstrap confidence interval as a subscript. Pre.\ GR is not in the bootstrap set, so it has no interval. The Off-policy rows use the common fixed $100$-step endpoint; the complete D4 Pareto metrics are recovered from the archived hardened evaluation.

\begin{table}[!h]\centering\setlength{\tabcolsep}{4pt}
\caption{D1, hardened protocol, including the task-adapted ANCE-PRF comparison. AUC / Attr / Same / Joint$_{[95\%\,\mathrm{CI}]}$ / Cond; densities in \% (AUC excepted). All rows use the same deterministic cases.}
\label{tab:appdc}
\begin{tabular}{llrrrrr}
\hline
Eval & Method & AUC & Attr & Same & Joint & Cond \\
\hline
Item & Avg. & 0.4793 & 12.63 & 90.51 & 8.29$_{[7.30,9.30]}$ & 87.65 \\
Item & Pre. GR & 0.4715 & 11.86 & 67.16 & 5.99 & 64.73 \\
Item & ANCE FirstP & 0.5853 & 29.38 & 52.87 & 10.81$_{[9.79,11.87]}$ & 36.81 \\
Item & PRF-$k0$ & 0.5853 & 29.38 & 52.87 & 10.81$_{[9.79,11.87]}$ & 36.81 \\
Item & ANCE-PRF & 0.5863 & 29.97 & 51.77 & 10.68$_{[9.66,11.73]}$ & 35.63 \\
Item & \mos{} & 0.5278 & 18.11 & 88.37 & 11.37$_{[10.10,12.70]}$ & 84.51 \\
Item & Off-policy & 0.5837 & 29.35 & 80.39 & 15.28$_{[13.86,16.75]}$ & 71.08 \\
Item & Adaptive & 0.5882 & 31.38 & 79.34 & 16.24$_{[14.70,17.70]}$ & 69.23 \\
Item & Pareto & 0.5992 & 35.34 & 76.29 & 17.36$_{[15.80,18.90]}$ & 65.90 \\
Item & Pareto-$\varepsilon$ & 0.6184 & 45.48 & 64.92 & \textbf{17.68}$_{[16.10,19.30]}$ & 54.22 \\
\hline
Pat. & Avg. & 0.4741 & 8.80 & 93.94 & 6.06$_{[5.30,7.00]}$ & 92.17 \\
Pat. & Pre. GR & 0.4659 & 8.58 & 73.30 & 4.58 & 68.70 \\
Pat. & ANCE FirstP & 0.5817 & 22.69 & 49.88 & 7.14$_{[6.40,7.91]}$ & 31.46 \\
Pat. & PRF-$k0$ & 0.5817 & 22.69 & 49.88 & 7.14$_{[6.40,7.91]}$ & 31.46 \\
Pat. & ANCE-PRF & 0.5826 & 23.22 & 48.45 & 7.03$_{[6.28,7.80]}$ & 30.26 \\
Pat. & \mos{} & 0.5171 & 11.55 & 89.92 & 7.74$_{[6.80,8.70]}$ & 83.92 \\
Pat. & Off-policy & 0.5776 & 20.81 & 84.50 & 11.77$_{[10.74,12.85]}$ & 75.74 \\
Pat. & Adaptive & 0.5831 & 22.68 & 83.15 & 12.71$_{[11.60,13.90]}$ & 73.78 \\
Pat. & Pareto & 0.5952 & 26.18 & 80.50 & 13.84$_{[12.70,15.10]}$ & 70.96 \\
Pat. & Pareto-$\varepsilon$ & 0.6144 & 36.57 & 69.61 & \textbf{14.97}$_{[13.80,16.20]}$ & 58.56 \\
\hline
\end{tabular}
\end{table}
The paired ANCE-PRF-minus-PRF-$k0$ \shortJoint{} differences are $-0.136$~pp (95\% CI $[-0.196,-0.076]$) for item holdout and $-0.110$~pp ($[-0.164,-0.060]$) for pattern holdout. ANCE-PRF exceeds Avg.\ and Pre.\ GR, falls below \mos{} by $0.69$~pp on item holdout and $0.71$~pp on pattern holdout, and remains substantially below \hppo{} on both splits.

\begin{table}[!h]\centering\setlength{\tabcolsep}{4pt}
\caption{D2, hardened protocol; densities in \%.}
\label{tab:appdp}
\begin{tabular}{llrrrrr}
\hline
Eval & Method & AUC & Attr & Same & Joint & Cond \\
\hline
Item & Avg. & 0.4977 & 3.89 & 78.75 & 2.81$_{[2.40,3.20]}$ & 73.82 \\
Item & Pre. GR & 0.4867 & 3.57 & 57.69 & 1.87 & 54.24 \\
Item & \mos{} & 0.5413 & 5.36 & 77.20 & 3.79$_{[3.20,4.50]}$ & 73.59 \\
Item & Off-policy & 0.5909 & 26.71 & 48.38 & 7.90$_{[6.74,9.16]}$ & 41.65 \\
Item & Adaptive & 0.5994 & 32.87 & 41.04 & 7.89$_{[6.70,9.20]}$ & 34.12 \\
Item & Pareto & 0.6068 & 22.47 & 52.68 & \textbf{10.28}$_{[8.90,11.70]}$ & 46.00 \\
Item & Pareto-$\varepsilon$ & 0.6007 & 35.23 & 37.80 & 8.82$_{[7.50,10.20]}$ & 31.43 \\
\hline
Pat. & Avg. & 0.5013 & 3.77 & 79.32 & 2.56$_{[2.20,3.00]}$ & 77.24 \\
Pat. & Pre. GR & 0.4886 & 3.66 & 58.11 & 1.98 & 60.70 \\
Pat. & \mos{} & 0.5426 & 7.16 & 75.94 & 4.99$_{[4.00,6.20]}$ & 72.49 \\
Pat. & Off-policy & 0.5965 & 33.28 & 45.23 & 11.15$_{[9.61,12.81]}$ & 41.94 \\
Pat. & Adaptive & 0.6035 & 38.88 & 36.85 & 10.42$_{[8.90,12.10]}$ & 34.12 \\
Pat. & Pareto & 0.6087 & 27.74 & 54.14 & \textbf{14.40}$_{[12.50,16.40]}$ & 48.64 \\
Pat. & Pareto-$\varepsilon$ & 0.6031 & 39.60 & 34.98 & 11.50$_{[9.80,13.30]}$ & 33.17 \\
\hline
\end{tabular}
\end{table}

\begin{table}[!h]\centering\setlength{\tabcolsep}{4pt}
\caption{D3, hardened protocol; densities in \%.}
\label{tab:appdu}
\begin{tabular}{llrrrrr}
\hline
Eval & Method & AUC & Attr & Same & Joint & Cond \\
\hline
Item & Avg. & 0.4565 & 8.97 & 93.17 & 8.03$_{[7.30,8.80]}$ & 92.56 \\
Item & Pre. GR & 0.4473 & 9.60 & 70.62 & 6.44 & 70.01 \\
Item & \mos{} & 0.5182 & 12.46 & 92.11 & 10.72$_{[9.70,11.80]}$ & 91.30 \\
Item & Off-policy & 0.5725 & 25.63 & 82.60 & 16.88$_{[15.24,18.58]}$ & 78.46 \\
Item & Adaptive & 0.5958 & 47.05 & 59.93 & 20.48$_{[18.60,22.40]}$ & 53.55 \\
Item & Pareto & 0.5916 & 42.08 & 66.97 & \textbf{21.42}$_{[19.60,23.30]}$ & 59.85 \\
Item & Pareto-$\varepsilon$ & 0.5941 & 45.45 & 62.58 & 21.20$_{[19.30,23.10]}$ & 55.66 \\
\hline
Pat. & Avg. & 0.4573 & 9.10 & 93.77 & 8.50$_{[7.90,9.10]}$ & 95.19 \\
Pat. & Pre. GR & 0.4525 & 9.14 & 73.56 & 6.75 & 75.86 \\
Pat. & \mos{} & 0.5054 & 11.77 & 91.06 & 10.72$_{[9.90,11.60]}$ & 92.74 \\
Pat. & Off-policy & 0.5600 & 20.46 & 84.78 & 15.45$_{[14.14,16.88]}$ & 81.79 \\
Pat. & Adaptive & 0.5845 & 39.93 & 58.45 & 16.10$_{[14.40,18.00]}$ & 47.79 \\
Pat. & Pareto & 0.5810 & 32.56 & 70.05 & \textbf{17.48}$_{[15.80,19.30]}$ & 59.53 \\
Pat. & Pareto-$\varepsilon$ & 0.5835 & 36.56 & 63.65 & 16.99$_{[15.20,18.90]}$ & 53.19 \\
\hline
\end{tabular}
\end{table}

\begin{table}[!h]\centering\setlength{\tabcolsep}{4pt}
\caption{D4, hardened protocol; densities in \%. The main table reports the complete Pareto-filtered procedure; Off-policy and Adaptive are mechanism ablations.}
\label{tab:appdb}
\begin{tabular}{llrrrrr}
\hline
Eval & Method & AUC & Attr & Same & Joint & Cond \\
\hline
Item & Avg. & 0.5631 & 16.21 & 87.53 & 9.53$_{[8.60,10.50]}$ & 82.93 \\
Item & Pre. GR & 0.5485 & 14.42 & 68.04 & 6.86 & 66.90 \\
Item & \mos{} & 0.6093 & 21.35 & 82.51 & 11.19$_{[10.20,12.20]}$ & 75.81 \\
Item & Off-policy & 0.6336 & 25.66 & 78.04 & \textbf{12.24}$_{[11.21,13.28]}$ & 69.89 \\
Item & Adaptive & 0.6466 & 28.96 & 72.17 & 12.22$_{[11.20,13.30]}$ & 62.92 \\
Item & Pareto & 0.6356 & 26.28 & 75.47 & 11.73$_{[10.69,12.79]}$ & 66.16 \\
Item & Pareto-$\varepsilon$ & 0.6353 & 26.10 & 76.05 & 11.81$_{[10.80,12.90]}$ & 66.99 \\
\hline
Pat. & Avg. & 0.5251 & 17.00 & 80.23 & 7.90$_{[7.40,8.50]}$ & 73.02 \\
Pat. & Pre. GR & 0.5117 & 14.69 & 66.44 & 5.82 & 62.13 \\
Pat. & \mos{} & 0.5871 & 22.67 & 73.66 & 8.99$_{[8.40,9.60]}$ & 65.16 \\
Pat. & Off-policy & 0.6216 & 28.11 & 67.70 & \textbf{9.12}$_{[8.56,9.67]}$ & 56.59 \\
Pat. & Adaptive & 0.6505 & 31.86 & 60.62 & 8.64$_{[8.10,9.20]}$ & 47.85 \\
Pat. & Pareto & 0.6337 & 28.78 & 64.32 & 8.64$_{[8.07,9.20]}$ & 51.95 \\
Pat. & Pareto-$\varepsilon$ & 0.6326 & 28.61 & 64.94 & 8.72$_{[8.20,9.30]}$ & 52.60 \\
\hline
\end{tabular}
\end{table}

\begin{figure}[!h]
\centering
\includegraphics[width=\columnwidth]{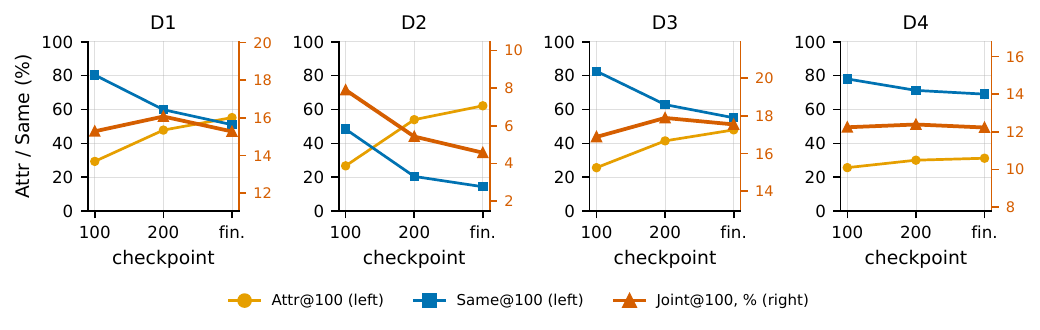}
\caption{Diagnostic off-policy checkpoint trajectory on the item split. As training continues (at $100\to200\to$final steps), \metricIssue{} (left axis) rises while \metricSame{} (left axis) falls, so the intersection \shortJoint{} (right axis, magnified) peaks early and then declines. D4 drifts least. This trajectory is not used to select the fixed-$100$ Off-policy control.}
\label{fig:drift}
\end{figure}

The off-policy checkpoints show the characteristic drift of unconstrained preference training (Figure~\ref{fig:drift}): raw attribute density keeps rising while same-pattern share falls and \shortJoint{} peaks early and then declines. This trajectory is a diagnostic motivation for reference anchoring, validation-controlled iteration, and the Pareto pair filter; the reported Off-policy control remains fixed at the common $100$-step endpoint.

\paragraph{The D4 exception.}\label{app:dbexception}
D4 is a failure case for strict Pareto filtering, not for \mos{} training overall. From the pretrained retriever to \mos{} SFT, \shortJoint{} rises from $6.86\%$ to $11.19\%$ on item holdout and from $5.82\%$ to $8.99\%$ on pattern holdout. At the common fixed $100$-step endpoint, Off-policy reaches $12.24\%$/$9.12\%$, whereas strict Pareto reaches $11.73\%$/$8.64\%$: paired differences of $-0.51$~pp $[-0.73,-0.29]$ and $-0.48$~pp $[-0.68,-0.29]$, respectively. Strict Pareto also has lower \metricSame{} on both splits ($75.5\%$ vs.\ $78.0\%$ item; $64.3\%$ vs.\ $67.7\%$ pattern), so the pair-level purity filter does not transfer to a globally Pareto-improved D4 policy.

Several diagnostics characterize the limited and unstable D4 preference signal. D4 is the smallest domain (Table~\ref{tab:data}: $478$K training rows and $500$ clusters); its \mos{} construction reuses each positive item about $31$ times on average, and SFT runs for two epochs. In round~1, only $2{,}415/46{,}000=5.25\%$ of training cases pass the top-1/top-2 reward-margin gate of $0.03$, versus $15.47\%$ on D1 and $14.79\%$ on D3; their mean/P90 margins are also smaller ($0.045/0.070$). The strict gate then retains $10{,}150/16{,}905=60.0\%$ of reward-separated pairs (versus $63.4$--$68.0\%$ on D1--D3), and D4's margin-qualified Joint winners have mean \metricSame{} $86.3\%$ (versus $92.4$--$95.0\%$), exposing a stronger conflict between reward and purity. D2 also has a low case-retention rate ($6.37\%$) yet benefits from the filter, so signal scarcity is not a sufficient explanation by itself. Candidate headroom is narrow on D4: in the pattern-holdout best-of-$N$ diagnostic, the full eight-candidate oracle over the HPPO policy reaches $8.60\%$, only $0.40$~pp above the SFT oracle ($8.20\%$; Table~\ref{tab:appbon}). Across three independent strict-Pareto runs, item \shortJoint{} remains below the fixed Off-policy control ($11.42$--$11.94\%$ vs.\ $12.24\%$), while the pattern ordering changes across seeds ($8.64$--$9.96\%$ vs.\ $9.12\%$; Table~\ref{tab:seed}), showing no stable Pareto advantage.

These observations jointly indicate that the strict filter removes signal in a D4 regime with small reward separations and limited candidate headroom. They do not identify data scale, reuse, margin retention, or candidate diversity as a unique cause without a single-factor intervention. We therefore restrict the Pareto-benefit claim to D1--D3 while reporting the same complete Pareto-filtered procedure in all four domains in Table~\ref{tab:main}.

Model-agnostic structural measurements over the training splits (frozen pretrained embeddings, annotation labels, train-side predicted density) distinguish D4 from the other domains. We project each sequence onto its dominant style axis---the first principal component of row-centered embeddings---and measure alignment with the positive--negative label direction. In D4 the two axes show stronger alignment (median $\cos^2 = 0.54$; $66\%$ of rows exceed $0.3$), whereas D1--D3 show weaker alignment (medians $0.08$--$0.10$; $16$--$20\%$ above threshold). Consistently, D4 positives concentrate at the style core: positive rate $91.6\%$ inside the median-distance shell versus $68.3\%$ in the periphery (gap $23.3$~pp; D1--D3 gaps $3.2$--$6.1$~pp), and predicted density correlates negatively with core distance (mean within-row correlation $-0.185$; D1--D3 near zero). These measurements indicate stronger geometric coupling of style and attribute on D4, consistent with less room for improving attribute density while preserving the seed pattern. The Pareto filter operates on retrieved same-pattern share, so these geometric correlations do not directly determine which candidate pairs it removes. The weaker alignment on D1--D3 is consistent with the larger observed in-pattern gains. This geometry is consistent with D4's lowest Joint-winner purity ($86.3\%$), narrow best-of-$N$ headroom ($0.40$~pp), and seed-unstable pattern ordering. We offer this as descriptive support rather than single-factor causal proof.

\section{Matched Three-Trigger Motivation Diagnostic}
\label{app:three-trigger}

\paragraph{Conclusion.}
Attribute-only and Pattern-only retrieval are both insufficient. On D1, a seed-independent high-density query attains $86.00\%$ \texttt{Attr@100} but almost zero \texttt{Same@100}, whereas an ordinary within-Pattern query preserves $89.82$--$94.30\%$ \texttt{Same@100} but has much lower \texttt{Attr@100}. Selecting the high-density tail within the seed Pattern gives the highest \texttt{Joint@100} on both item and complete-Pattern holdouts (Table~\ref{tab:three-trigger}). This matched result directly substantiates the motivating tradeoff: the desired query must combine Attribute-oriented movement with seed-Pattern conditioning.

\paragraph{Controlled construction.}
We use the frozen D1 evaluation protocol with $500$ cases in each holdout. Every query is the centroid of exactly $50$ frozen $1{,}536$-dimensional item embeddings; only the identities of those items change. \emph{Pattern random-50} deterministically samples $50$ items from the first $150$ lower-density items in the evaluation case's metric-ordered, same-Pattern trajectory. It retains seed-Pattern conditioning but does not explicitly target a high-Attribute region. \emph{Global high-density-50} first scans the strict D1 training split, takes the $50$ highest-predicted-density unique items within each training trajectory, and selects the coherent same-Pattern set with the highest mean predicted density. Its one fixed centroid is then reused for every evaluation case. The selected set has its own training Pattern, but the resulting query is independent of the current case's seed Pattern; it is therefore an Attribute-oriented control rather than the production classifier itself. \emph{Tail-50} averages the final $50$ high-predicted-density items in the current evaluation case's same-Pattern trajectory, combining Pattern conditioning and Attribute-oriented selection.

All three queries use the same cases, frozen ANN index, and evaluator. Retrieval requests the Top-$200$, excludes that arm's $50$ trigger items, deduplicates the results, and retains the Top-$100$. \texttt{Attr@100}, \texttt{Same@100}, and \texttt{Joint@100} are computed on the retrieved items. Uncertainty uses $10{,}000$ paired bootstrap draws over input-Pattern blocks.

\begin{table}[!h]
\centering
\setlength{\tabcolsep}{4pt}
\caption{Matched D1 three-trigger diagnostic. Metric entries are percentages and differences are percentage points; each holdout contains $500$ cases. Intervals use $10{,}000$ paired input-Pattern-block bootstrap draws.}
\label{tab:three-trigger}
\begin{tabular}{llrrrr}
\hline
Holdout & Query & Attr@100 & Same@100 & Joint@100 & \shortstack{Tail$-$Random\\$\Delta$Joint [95\% CI]} \\
\hline
Item & Pattern random-50 & 7.82 & \textbf{89.82} & 6.74 & -- \\
Item & Global high-density-50 & \textbf{86.00} & 0.11 & 0.10 & -- \\
Item & Tail-50 & 10.62 & 88.73 & \textbf{8.56} & $\mathbf{+1.82}\ [\mathbf{+1.36},\mathbf{+2.28}]$ \\
\shortstack[l]{Complete\\Pattern} & Pattern random-50 & 4.90 & \textbf{94.30} & 4.20 & -- \\
\shortstack[l]{Complete\\Pattern} & Global high-density-50 & \textbf{86.00} & 0.01 & 0.01 & -- \\
\shortstack[l]{Complete\\Pattern} & Tail-50 & 7.33 & 92.76 & \textbf{6.01} & $\mathbf{+1.81}\ [\mathbf{+1.18},\mathbf{+2.51}]$ \\
\hline
\end{tabular}
\end{table}

The repeated $86.00\%$ \texttt{Attr@100} is expected because the fixed global query retrieves the same Top-$100$ list in both evaluation cohorts; \texttt{Same@100} and \texttt{Joint@100} are recomputed against each case's seed Pattern. Tail-50 directly uses the evaluation case's ordered tail and is therefore an oracle target-geometry diagnostic, not a deployable retriever. The learned-method results are evaluated separately in Table~\ref{tab:main}, where AvgPool $<$ \mos{} $<$ \hppo{} in \shortJoint{} on both D1 holdouts.

\section{Best-of-N Oracle, Candidate, Flow-Surrogate, and Guidance Analyses}
\label{app:bon-analyses}

\paragraph{Best-of-N oracle (Table~\ref{tab:appbon}).} For a fixed policy we score all eight candidates with true online \shortJoint{} and compare the served single query against per-case oracles (best static candidate, best policy candidate, best of all eight). On D1--D3 and on the D4 item split, the HPPO single query exceeds the SFT eight-candidate oracle; an oracle over the HPPO candidates adds a further $15$--$27\%$ relative on D1--D3, indicating headroom for an inference-time reranker. D4 is the exception on the pattern split, with small generation--selection headroom on both splits.

\begin{table}[!h]\centering\setlength{\tabcolsep}{4pt}
\caption{Best-of-$N$ oracle on $1{,}000$ held-out cases/split (\shortJoint{} in \%). ``or'' = per-case oracle. The two static-or columns coincide by construction, as the static tail centroids do not depend on the policy.}
\label{tab:appbon}
\begin{tabular}{llrrrrrr}
\hline
Eval & Dom & \shortstack{SFT\\static-or} & \shortstack{SFT\\8-or} & \shortstack{HPPO\\single} & \shortstack{HPPO\\static-or} & \shortstack{HPPO\\gr-or} & \shortstack{HPPO\\8-or} \\
\hline
Item & D1 & 8.70 & 10.20 & \textbf{12.60} & 8.70 & 15.30 & 15.70 \\
Pat. & D1 & 9.30 & 10.50 & \textbf{13.50} & 9.30 & 16.40 & 17.10 \\
\hline
Item & D2 & 1.60 & 1.90 & \textbf{5.60} & 1.60 & 6.50 & 6.80 \\
Pat. & D2 & 1.30 & 1.70 & \textbf{8.00} & 1.30 & 9.10 & 9.20 \\
\hline
Item & D3 & 8.80 & 9.70 & \textbf{15.70} & 8.80 & 18.00 & 19.10 \\
Pat. & D3 & 6.20 & 8.00 & \textbf{12.80} & 6.20 & 15.60 & 15.90 \\
\hline
Item & D4 & 10.90 & 11.90 & 12.60 & 10.90 & 13.40 & 13.70 \\
Pat. & D4 & 7.70 & 8.20 & 7.20 & 7.70 & 7.70 & 8.60 \\
\hline
\end{tabular}
\end{table}

\paragraph{Candidate-policy ablation (D1, off-policy regime, common $400$-step endpoint).} Table~\ref{tab:cand} reports the per-pool numbers; both candidate families contribute.

\begin{figure}[!h]
\centering
\includegraphics[width=\columnwidth]{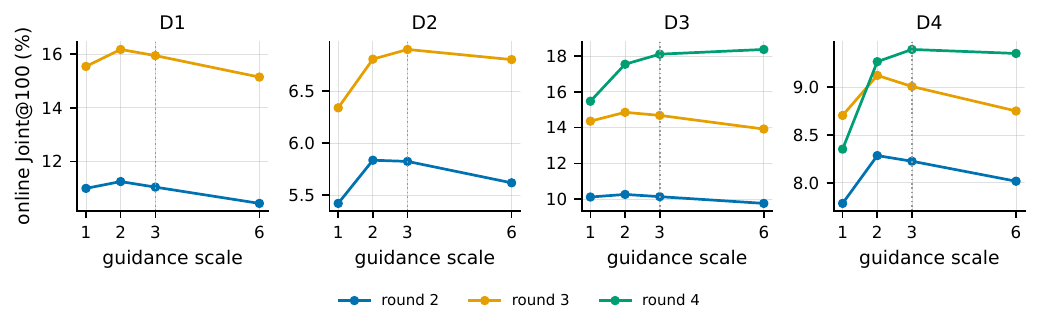}
\caption{Guidance-fan calibration: mean true online \shortJoint{} of the policy candidate at each guidance scale, by domain and preference round. The dotted line marks the serving default ($s{=}3$). Scales $2$--$3$ are consistently near the peak.}
\label{fig:guidance}
\end{figure}

\paragraph{Guidance-scale calibration (Figure~\ref{fig:guidance}).} Mean true online \shortJoint{} per fan scale, by domain and preference round. Scales $2$--$3$ generally perform near the peak, while the best scale varies by domain and round; scale $1$ is not always the worst. The serving default $3.0$ is near-optimal.

\paragraph{Hyperparameter sensitivity (D1).} We sweep the two main \hppo{} hyperparameters---the reward margin $\delta$ and the preference temperature $\beta$---around their defaults ($\delta=0.03$, $\beta=0.05$) on D1, holding everything else fixed and reusing the cached round-one labels. Table~\ref{tab:hparam} reports the primary \shortJoint{} metric under the hardened protocol. It is stable across both knobs (within about one percentage point), and the defaults sit at or near the optimum, so the reported gains do not depend on hyperparameter tuning. These are single-domain runs, so absolute values are not directly comparable to the main tables.

\begin{table}[!t]
\centering
\caption{\hppo{} hyperparameter sensitivity on D1: primary \shortJoint{} metric (in percent) under the hardened protocol, varying the reward margin $\delta$ and the preference temperature $\beta$ around their defaults. Single-domain runs; absolute values are not directly comparable to the main tables.}
\label{tab:hparam}
\begin{tabular}{llrr}
\hline
Hyperparameter & Value & Item & Pattern \\
\hline
Margin $\delta$ & 0.01 & 17.82 & 15.77 \\
 & 0.03 (default) & \textbf{17.99} & \textbf{15.94} \\
 & 0.05 & 17.21 & 15.01 \\
\hline
Temperature $\beta$ & 0.025 & 17.13 & 15.30 \\
 & 0.05 (default) & \textbf{17.99} & \textbf{15.94} \\
 & 0.10 & 17.54 & 15.47 \\
\hline
\end{tabular}
\end{table}

\paragraph{Retrieval-depth sensitivity (D1).} To isolate serving/evaluation depth from retraining effects, we keep the checkpoints trained with the default $K=100$ supervision fixed. On each D1 item- and pattern-holdout query, we retrieve one ranked top-200 ANN list and evaluate its exact prefixes at $K\in\{20,50,100,200\}$. We use the same hardened sampling protocol (500 cluster/score-stratified cases per split and 20 generated queries per case) and paired case-level bootstrap with $10{,}000$ resamples. All depths therefore share cases, generated queries, ANN rankings, and scorer outputs; the relevant comparisons are within this matched sweep rather than against separately executed online runs. Table~\ref{tab:retrieval-k} shows that \hppo{} retains the highest \shortJoint{} in all eight split--depth settings, followed by \mos{} and Avg. The \hppo{}--\mos{} \shortJoint{} margin ranges from $4.04$ to $8.58$ pp on item holdout and from $4.37$ to $8.79$ pp on pattern holdout; all eight paired $95\%$ confidence intervals are strictly above zero. Absolute densities decrease as deeper, lower-ranked neighbors enter the denominator, but the method ordering is unchanged. This controlled sweep establishes robustness of the reported ranking to evaluation depth; it does not claim robustness to rebuilding the $K$-dependent labels/rewards and retraining at every $K$.

\begin{table}[!t]
\centering
\setlength{\tabcolsep}{5pt}
\caption{D1 retrieval-depth sensitivity with fixed $K{=}100$-trained checkpoints. Metrics are Attr, Same, and Joint (in percent); smaller depths are exact prefixes of the same ranked top-200 ANN result.}
\label{tab:retrieval-k}
\begin{tabular}{lllrrr}
\hline
Eval & $K$ & Method & Attr & Same & Joint \\
\hline
Item & 20 & Avg. & 11.05 & 94.04 & 10.14 \\
Item & 20 & \mos{} & 16.55 & 92.21 & 14.66 \\
Item & 20 & \hppo{} & 33.79 & 79.74 & \textbf{23.24} \\
\hline
Item & 50 & Avg. & 10.02 & 92.12 & 8.93 \\
Item & 50 & \mos{} & 14.70 & 90.06 & 12.42 \\
Item & 50 & \hppo{} & 29.93 & 77.98 & \textbf{19.06} \\
\hline
Item & 100 & Avg. & 9.13 & 90.31 & 7.86 \\
Item & 100 & \mos{} & 13.20 & 88.05 & 10.66 \\
Item & 100 & \hppo{} & 26.79 & 76.04 & \textbf{15.86} \\
\hline
Item & 200 & Avg. & 8.38 & 87.88 & 6.91 \\
Item & 200 & \mos{} & 11.69 & 85.66 & 8.96 \\
Item & 200 & \hppo{} & 23.85 & 73.92 & \textbf{12.99} \\
\hline
Pat. & 20 & Avg. & 6.13 & 96.03 & 5.43 \\
Pat. & 20 & \mos{} & 9.16 & 92.87 & 8.02 \\
Pat. & 20 & \hppo{} & 23.98 & 82.47 & \textbf{16.81} \\
\hline
Pat. & 50 & Avg. & 6.01 & 95.26 & 5.26 \\
Pat. & 50 & \mos{} & 8.50 & 91.71 & 7.19 \\
Pat. & 50 & \hppo{} & 21.80 & 81.16 & \textbf{14.60} \\
\hline
Pat. & 100 & Avg. & 5.90 & 94.42 & 5.12 \\
Pat. & 100 & \mos{} & 8.13 & 90.77 & 6.79 \\
Pat. & 100 & \hppo{} & 19.60 & 80.42 & \textbf{12.51} \\
\hline
Pat. & 200 & Avg. & 5.69 & 93.28 & 4.87 \\
Pat. & 200 & \mos{} & 7.58 & 89.50 & 6.14 \\
Pat. & 200 & \hppo{} & 17.40 & 79.60 & \textbf{10.51} \\
\hline
\end{tabular}
\end{table}

\paragraph{Flow-surrogate reliability diagnostics.} We evaluate the actual reference-relative score used by \hppo{} on the D1 item and pattern holdouts, with 1{,}000 cases, eight fixed candidates per case, their cached true online labels, and 64 independent score draws per candidate and checkpoint. The policy is the Pareto checkpoint and the reference is its SFT initialization. Pair accuracy uses the 64-draw mean and case-level bootstrap; Flip@1/8 compares a one/eight-draw estimate from the first 32 draws against an independent consensus from the remaining 32. This repeated-draw protocol diagnoses estimator stability and does not change the one-draw training estimator. The same reward-gap trend replicates with the Adaptive checkpoint: item accuracy is $71.94/79.15/88.48\%$ and pattern accuracy is $63.69/67.99/82.87\%$ over the three increasing gap buckets in Table~\ref{tab:surrogate-main}.

Table~\ref{tab:surrogate-var} directly tests the non-policy candidate family used by \hppo{}. Static tail centroids are not sampled from the generative policy, yet their score variance is lower than that of policy-generated candidates on both splits.

\begin{table}[!h]
\centering
\setlength{\tabcolsep}{5pt}
\caption{Candidate-level Monte Carlo stability across 64 repeated evaluations of each fixed candidate. Relative SD is normalized by the policy-generated row within each split.}
\label{tab:surrogate-var}
\begin{tabular}{llrrr}
\hline
Eval & Candidate source & Candidates & Mean score SD & Relative SD \\
\hline
Item & Static (not policy-sampled) & 4{,}000 & $5.5\times10^{-5}$ & $0.76\times$ \\
Item & Policy-generated & 4{,}000 & $7.2\times10^{-5}$ & $1.00\times$ \\
Pattern & Static (not policy-sampled) & 4{,}000 & $6.0\times10^{-5}$ & $0.81\times$ \\
Pattern & Policy-generated & 4{,}000 & $7.4\times10^{-5}$ & $1.00\times$ \\
\hline
\end{tabular}
\end{table}

\section{Pattern-Proxy and Metric-Predictor Validation}
\label{app:patternproxy}

\subsection{External Semantic Validation of the Pattern Proxy}

We evaluate the paper's frozen multimodal embeddings and fixed issue-specific $k$-means centers on a human-annotated pattern set whose labels were produced independently of the proxy construction and were not derived from the cluster assignments. Before any proxy assignment or agreement computation, we freeze the human-pattern inventory: item IDs are deduplicated within each pattern, only positive annotations with valid embeddings are retained, patterns with fewer than $10$ valid items are excluded, and patterns larger than $600$ are sampled without replacement using fixed per-pattern seeds. No cluster ID, center distance, or agreement result is used for filtering. The resulting evaluation set contains $66$ human patterns, $15{,}754$ pattern--item memberships, and $15{,}414$ unique annotated items ($238.7$ items per pattern on average; median $173$).

Let $H$ denote that a pair shares at least one observed human pattern and $C$ that it maps to the same proxy cluster. We measure both forward concentration, $P(C\mid H)$, and reverse annotated-subset purity, $P(H\mid C,\text{ both annotated})$. For the forward null, we permute fixed proxy assignments within predefined evaluation strata while preserving cluster occupancy, human-pattern sizes, and complete multi-label sets. For the reverse baseline, we compare with random-pair label agreement in the same stratum, weighted by its observed same-center pair count. All matched comparisons and pooling remain within these strata so that missing labels are not treated as negative annotations.

\begin{table}[!h]
\centering
\setlength{\tabcolsep}{5pt}
\caption{External semantic validation of the frozen pattern proxy. Obs.\ and random rates and confidence intervals are percentages. Forward uncertainty bootstraps human patterns and its significance uses $2{,}000$ matched permutations. Reverse uncertainty resamples collision centers within predefined evaluation strata before pooling; same-center pairs within a center are combinatorial, not independent observations.}
\label{tab:pattern-proxy-validation}
\begin{tabular}{l>{\raggedright\arraybackslash}p{0.26\linewidth}rrrr}
\hline
Direction & Statistical support & Obs. & \shortstack{Matched\\random} & Lift & 95\% CI \\
\hline
Forward $P(C\mid H)$ & $66$ patterns; $15{,}754$ memberships & 18.19 & 2.87 & $6.33\times$ & $[13.78,23.45]$ \\
Reverse $P(H\mid C)$ & $65$ patterns; $1{,}109$ centers; $709{,}661$ pairs & 80.00 & 10.89 & $7.34\times$ & $[67.30,88.34]$ \\
\hline
\end{tabular}
\end{table}

The forward enrichment is significant at the resolution of the permutation test ($p=1/2001<0.0005$). The reverse numerator contains $567{,}745$ agreeing same-center pairs. One single-pattern evaluation stratum is omitted from the reverse statistic because both its observed and random-pair agreement are mechanically $100\%$.

These two directions answer complementary questions: human-similar items concentrate in the proxy, and same-proxy annotated items are substantially more likely to share a human pattern. Absolute agreement need not approach $100\%$ because the $500$/$1{,}000$ issue-specific proxy centers are substantially finer-grained than the human annotations; one human-recognizable pattern may contain several valid local embedding neighborhoods. We therefore interpret the proxy as strongly aligned with human semantics but not as a one-to-one human taxonomy.

\subsection{Cluster-Count Sensitivity and Deployment Guidance}
\label{app:c-sensitivity}

We isolate the pattern-cluster count on D1 with $C\in\{500,1000,2000\}$. The three settings share the item universe, frozen embeddings, item exclusions, human-pattern annotations, model initialization, training objectives and hyperparameters, and evaluation protocol. We rebuild every cluster-dependent assignment, split, sequence, and model for $C=500$ and $C=2000$; the frozen $C=1000$ reference is re-evaluated with the same evaluator and case protocol. Auxiliary choices are fixed before evaluation, so held-out \shortJoint{} is not used to select $C$.

On the same $36$ independently curated D1 human patterns ($9{,}225$ memberships; $8{,}900$ unique items), Table~\ref{tab:c-semantic} evaluates the two directions defined above. Forward concentration measures how often pairs from one human pattern remain together; reverse purity measures how often same-cluster annotated pairs share a human pattern.

\begin{table}[!h]
\centering
\setlength{\tabcolsep}{4pt}
\caption{D1 semantic granularity versus pattern-cluster count. Forward and reverse entries are percentages with $95\%$ confidence intervals; the forward null uses $2{,}000$ matched permutations. Pairs is the number of same-center annotated pairs.}
\label{tab:c-semantic}
\begin{tabular}{rrrrrr}
\hline
$C$ & Forward [95\% CI] & Matched null & Lift & Reverse [95\% CI] & Pairs \\
\hline
500  & 21.09 [15.58, 27.43] & 2.04 & $10.36\times$ & 67.82 [56.29, 82.08] & 590{,}736 \\
1000 & 18.44 [12.24, 26.01] & 1.53 & $12.02\times$ & 82.80 [68.12, 91.14] & 482{,}572 \\
2000 & 14.55 [9.89, 20.19]  & 1.11 & $13.14\times$ & 84.49 [73.08, 91.76] & 334{,}652 \\
\hline
\end{tabular}%
\end{table}

All three settings exceed their matched forward nulls ($p=1/2001$). The coarser $C=500$ partition retains more within-pattern pairs but has $15.0$ percentage points lower reverse purity than $C=1000$, consistent with merging distinct human patterns. Increasing to $C=2000$ adds only $1.7$ points of reverse purity over $C=1000$ while losing $3.9$ points of forward concentration and $31\%$ of same-center annotated-pair support, consistent with additional fragmentation.

We next rebuild the complete $C$-dependent training and strict item/pattern-holdout evaluation pipeline. Table~\ref{tab:c-retrieval} reports paired differences on the primary \shortJoint{} metric; each cell averages generated queries within each of the same $500$ cases and uses $10{,}000$ paired case-bootstrap resamples. Reporting within-$C$ differences is essential because changing $C$ changes both the same-pattern target and the pattern-holdout split, so raw \shortJoint{} values are not a common target across cluster counts.

\begin{table}[!h]
\centering
\setlength{\tabcolsep}{5pt}
\caption{D1 cluster-count sensitivity of the method ordering. Entries are paired \shortJoint{} differences in percentage points with $95\%$ confidence intervals. Comparisons are within each fixed $C$ and split.}
\label{tab:c-retrieval}
\begin{tabular}{llrr}
\hline
$C$ & Eval & \mos{} $-$ Avg. & \hppo{} $-$ \mos{} \\
\hline
500  & Item    & 2.77 [2.21, 3.37] & 7.80 [6.81, 8.81] \\
     & Pattern & 2.56 [1.96, 3.23] & 6.83 [5.93, 7.76] \\
1000 & Item    & 2.66 [2.14, 3.19] & 4.55 [3.85, 5.26] \\
     & Pattern & 1.39 [1.02, 1.77] & 5.26 [4.65, 5.88] \\
2000 & Item    & 1.55 [1.15, 2.00] & 4.37 [3.76, 5.01] \\
     & Pattern & 1.69 [1.33, 2.10] & 4.34 [3.73, 4.98] \\
\hline
\end{tabular}
\end{table}

The ordering Avg.\ $<$ \mos{} $<$ \hppo{} holds in all six cluster-count--split cells, and every paired interval excludes zero. The method improvements therefore persist across the tested granularities rather than depending on the default $C=1000$.

For deployment, we first require significant forward enrichment over its matched null and inspect reverse purity, then choose the elbow where reverse purity has nearly saturated while forward concentration, cluster occupancy, and sequence-construction support remain sufficient. We freeze this training-side choice before held-out evaluation. These diagnostics select $C=1000$ for D1: $C=500$ is on the merging side of the tradeoff, whereas $C=2000$ provides little additional reverse purity while reducing forward concentration and same-center annotated-pair support. The same diagnostic procedure can be applied independently to each domain; the selected cluster count remains domain-specific.

\subsection{Metric-Predictor Validation}

Super-pattern construction is performed independently within each applicable domain. We first $L_2$-normalize every eligible labeled-anchor embedding, average the embeddings assigned to each fine cluster, and run MiniBatchKMeans on these fine-cluster centroids (seed $20260516$, \texttt{n\_init}$=10$). This produces $100$ families for D1 and $50$ families each for D2 and D3. The grouping is formed before the holdout and uses only frozen embeddings, not density targets. Pattern-out maps fine-cluster IDs to a deterministic $20\%$ hash bucket and removes every anchor in the selected clusters from predictor fitting. Super-pattern-out applies the same rule to family IDs and removes every anchor in every member cluster. The percentage therefore refers to group-ID hash buckets, not necessarily $20\%$ of rows. D4 uses a separately available held-out-cluster validation and has no super-pattern split. These predictor diagnostics are distinct from the $5\%$ pattern holdouts used in the main retrieval experiments.

Ridge regression is intentionally a low-capacity probe: \mos{} construction requires a useful ordinal signal, not exact density calibration or a nonlinear final decision boundary. Table~\ref{tab:density} tests this requirement directly. Across the available strict pattern-out and super-pattern-out cells, AUC20/80 remains $0.882$--$0.973$, Spearman correlation remains $0.530$--$0.752$, and top-to-bottom-decile true-density lift remains $8.67$--$25.71\times$. These results show that the ranking signal generalizes beyond memorized cluster identity. The downstream intervention agrees: ascending order beats random and descending order in all eight item- and pattern-holdout domain cells in Table~\ref{tab:order}.

\begin{table}[!h]\centering\setlength{\tabcolsep}{4pt}
\caption{Metric (attribute-density) predictor validation. AUC20/80 evaluates whether the predictor separates the lowest-density 20\% and highest-density 20\% anchors. Lift is Top10 true density divided by Bottom10 true density. D4 is shown on item-hash and pattern-out only; no super-pattern-out split is available for D4.}
\label{tab:density}
\begin{tabular}{l>{\raggedright\arraybackslash}p{0.22\linewidth}rrrrr}
\hline
Domain & Split & AUC20/80 & Spearman & \shortstack{Top10\\(\%)} & \shortstack{Bottom10\\(\%)} & Lift \\
\hline
D1 & item-hash & 0.959 & 0.663 & 39.40 & 1.75 & 22.50 \\
D1 & pattern-out & 0.929 & 0.606 & 35.60 & 2.26 & 15.78 \\
D1 & super-pattern-out & 0.900 & 0.552 & 25.00 & 1.95 & 12.83 \\
D2 & item-hash & 0.945 & 0.647 & 19.90 & 1.07 & 18.53 \\
D2 & pattern-out & 0.951 & 0.661 & 21.00 & 0.99 & 21.10 \\
D2 & super-pattern-out & 0.920 & 0.591 & 15.50 & 1.29 & 12.01 \\
D3 & item-hash & 0.957 & 0.659 & 45.40 & 3.18 & 14.28 \\
D3 & pattern-out & 0.934 & 0.621 & 40.90 & 3.96 & 10.33 \\
D3 & super-pattern-out & 0.882 & 0.530 & 28.60 & 3.30 & 8.67 \\
D4 & item-hash & 0.993 & 0.800 & 58.50 & 1.21 & 48.38 \\
D4 & pattern-out & 0.973 & 0.752 & 42.90 & 1.67 & 25.71 \\
\hline
\end{tabular}
\end{table}

\section{Original Full-Protocol Results (single-domain instantiation)}
\label{app:original-results}

These are the earlier formal results under a full (non-strict, larger-case) evaluation of the single-domain instantiation. They agree with the strict-protocol conclusions but are not directly comparable in magnitude to the hardened tables. The stage names match the main tables, except that CPT here is the per-domain continuation pretraining of this single-domain instantiation rather than the shared multi-domain CPT.

\paragraph{Headline.} \mos{} CPT/SFT improves \shortJoint{} over the pretrained retriever on all four domains: D1 $5.07\!\to\!10.57$, D2 $4.90\!\to\!7.18$, D3 $6.37\!\to\!12.05$, D4 $6.49\!\to\!9.41$ (all \%; $+44.9\%$ to $+108.4\%$ relative); per-stage detail for D2 and D3 is in Table~\ref{tab:apporig}.

\begin{table}[!h]\centering\setlength{\tabcolsep}{5pt}
\caption{D2 and D3, full-protocol single-domain instantiation (AUC; Attr / Cond / Joint / Same in \%).}
\label{tab:apporig}
\begin{tabular}{lrrrrr}
\hline
Model & AUC & Attr & Cond & Joint & Same \\
\hline
D2 Pre.\ GR & 0.4898 & 9.53 & 60.45 & 4.90 & 62.02 \\
D2 CPT & 0.5026 & 9.68 & 65.46 & 5.30 & 64.89 \\
D2 SFT & 0.5439 & 11.24 & 77.70 & \textbf{7.18} & 81.84 \\
D3 Pre.\ GR & 0.4457 & 9.27 & 68.42 & 6.37 & 69.67 \\
D3 CPT & 0.4727 & 9.89 & 72.04 & 7.08 & 73.33 \\
D3 SFT & 0.5150 & 12.76 & 94.82 & \textbf{12.05} & 97.12 \\
\hline
\end{tabular}
\end{table}

\paragraph{Target design (D4, full protocol).} The oracle tail centroid reaches \shortJoint{} $9.98\%$ at conditional purity $95.53\%$; oracle top-$1$/top-$5$ targets reach higher attribute density ($11.80$/$12.12$) but lower \shortJoint{} ($6.34$/$6.94$) because conditional purity collapses ($70.59$/$72.96$). The learned \mos{} model reaches \shortJoint{} $9.41\%$ at the highest conditional purity ($99.54\%$). All values are percentages. The tail centroid is therefore the right supervision target: sharper top-$k$ targets trade the intersection away (Table~\ref{tab:appbfs}).

\begin{table}[!h]\centering\setlength{\tabcolsep}{5pt}
\caption{D4, full-protocol single-domain instantiation, and target design (AUC; Attr / Cond / Joint / Same in \%).}
\label{tab:appbfs}
\begin{tabular}{lrrrrr}
\hline
Model & AUC & Attr & Cond & Joint & Same \\
\hline
Pre.\ GR & 0.5129 & 7.60 & -- & 6.49 & -- \\
\mos{} (SFT) & 0.5731 & 9.47 & 99.54 & \textbf{9.41} & 98.20 \\
target centroid & 0.6498 & 10.74 & 95.53 & 9.98 & 96.86 \\
target top-1 & 0.6217 & 11.80 & 70.59 & 6.34 & 66.06 \\
target top-5 & 0.6681 & 12.12 & 72.96 & 6.94 & 77.45 \\
\hline
\end{tabular}
\end{table}

\section{Public Semi-Truths Benchmark}
\label{app:public-semitruths}

\paragraph{Dataset and crossed-axis construction.} Semi-Truths contains $27{,}600$ real source images and about $1.33$ million targeted AI-augmented images with source and edit metadata \citep{pal2024semitruths}. We use frozen 768-dimensional CLIP ViT-L/14 embeddings and a fixed QC-passed $98{,}481$-image retrieval bank. The Attribute is supplied by public provenance: original, unedited images are positive and edited descendants are negative. To construct a content Pattern without entangling it with edit status, we fit $C=128$ spherical clusters on original source images only; every edited descendant inherits the Pattern of its source family. The public images, source-family links, centers, and assignments are inspectable, while the clusters remain an embedding proxy rather than a claimed one-to-one human taxonomy.

\paragraph{Sequences, training, and holdouts.} The frozen construction contains $1{,}455{,}360$ training and $24{,}072$ validation sequences. Each \mos{} SFT case presents 12 same-Pattern items and targets the centroid of a four-item higher-original-density Tail. Before SFT, label-free source-family multi-view pretraining provides a continuous-retrieval prior and teaches the public 768-dimensional generator to condition on related visual inputs; it does not use original/edited labels. Tail-centroid SFT then maps the same-Pattern prefix toward a high-Attribute local region rather than a potentially noisy single item. The resulting pattern-preserving improvement is supported on the item holdout, which contains unseen items from Patterns represented during training; we do not extend this stability claim to the six completely unseen Patterns. Finally, \hppo{} closes the remaining proxy-to-objective gap with exact train-bank \shortJoint{} labels, a hybrid static/policy candidate pool, and the \metricSame{}-preserving Pareto pair filter. Source families are atomic across all splits. The final item holdout has 272 cases from 17 represented Patterns; the complete-Pattern holdout has 96 cases from six Patterns absent from every weight update, including pretraining, SFT, preference optimization, and reward labeling. Checkpoints are selected on disjoint development splits before either final split is opened.

\begin{table}[!h]
\centering
\setlength{\tabcolsep}{5pt}
\caption{Public Semi-Truths exact Top-100 retrieval. Attr, Same, and Joint are percentages. The Pattern split holds out six complete Patterns from every weight-updating stage.}
\label{tab:public-semitruths}
\begin{tabular}{llrrr}
\hline
Eval & Method & Attr & Same & Joint \\
\hline
Item & Avg. & 46.93 & 80.68 & 39.14 \\
Item & Tail oracle & 56.20 & 74.65 & 41.12 \\
Item & SFT & 60.33 & 78.49 & 46.61 \\
Item & \hppo{} & 65.56 & 78.39 & \textbf{50.30} \\
\hline
Pattern & Avg. & 50.60 & 61.42 & 30.70 \\
Pattern & Tail oracle & 57.81 & 58.07 & 32.12 \\
Pattern & SFT & 67.86 & 45.33 & 31.28 \\
Pattern & \hppo{} & 71.95 & 45.44 & \textbf{33.21} \\
\hline
\end{tabular}
\end{table}

The item split exhibits the intended mechanism: SFT and \hppo{} retain nearly all of Avg.'s \metricSame{} while increasing Attribute and \shortJoint{}. On both final splits, \hppo{} raises Attribute over SFT with \metricSame{} essentially unchanged, so its \shortJoint{} gain does not come from additional Pattern drift. The primary point ordering Avg. $<$ SFT $<$ \hppo{} holds on both splits. With $2{,}000$ paired bootstrap draws over Pattern blocks, the item-holdout SFT-minus-Avg. gain is $+7.47$~pp (95\% CI $[+5.11,+9.74]$), and \hppo{} adds $+3.69$~pp over SFT ($[+2.85,+4.60]$). On the complete-Pattern holdout, \hppo{} adds $+1.93$~pp over SFT ($[+1.20,+2.81]$). The SFT-minus-Avg. point estimate there is $+0.58$~pp, but its six-Pattern-block interval $[-3.51,+3.58]$ includes zero; we therefore claim strict point ordering and a statistically resolved \hppo{} increment, not a resolved SFT--Avg. difference on complete unseen Patterns.

\section{Negative and Diagnostic Results}

We record results that did not enter the main paper, both for completeness and to motivate the final recipe.

\paragraph{Unconstrained off-policy preference (negative).} An earlier $200$k-case off-policy/online preference suite without the Pareto filter and with longer training was \emph{below} the SFT initialization on \shortJoint{} on both holdouts; hybrid/online variants raised attribute density and AUC but collapsed conditional purity (reward hacking toward global attribute retrieval). This motivated the fixed SFT reference, validation-controlled iteration, and the Pareto pair filter; it is separate from the fixed-$100$ Off-policy mechanism control in Table~\ref{tab:hppo-abl}.

\paragraph{Local/proxy reward (negative).} Pure-listwise training against a local approximate-retrieval bank, and a selected-case proxy bank, were unreliable: local reward over-amplified some candidates (e.g.\ Rocchio-style vectors) relative to true online retrieval, and downstream metrics degraded. True online \shortJoint{} labels were necessary.

\paragraph{Guidance fan is necessary for the filter (negative).} Replacing the guidance fan with duplicate policy samples (no diversity) under the Pareto filter produced validation scores below baseline and early-stopped at SFT: the candidate diversity of the fan is what makes the pair filter effective.

\paragraph{The eps knob is not a clean curve (diagnostic).} Sweeping the purity-tolerance $\varepsilon$ to trace an attribute--purity frontier does not yield a clean monotone curve, because each run early-stops at a different round and the round count is itself a frontier slider; $\varepsilon$ and training amount are entangled. We therefore report only the strict ($\varepsilon{=}0$, ``Pareto'') and $\varepsilon{=}0.02$ (``Pareto-$\varepsilon$'') settings rather than a frontier; a tradeoff-parameter-conditioned policy is the proper route to a controllable frontier and is left to future work.

\section{Reproducibility and Anonymization}

The four domains are anonymized because they use our internal data. The method depends only on: frozen item embeddings; a binary or real-valued attribute scorer; a vector index for nearest-neighbor retrieval; and sparse online recall-density labels for predictor training---all common components of large-scale item-embedding retrieval systems. \mos{} construction, \mos{} CPT/SFT, \hppo{}, and the evaluation protocol are described in the main text and Appendices~A--B. All reported numbers derive from fixed deterministic case samples; the significance analysis covers evaluation-sampling variance via paired bootstrap and training-run variability via three independent re-runs of the preference stage with different random seeds (Table~\ref{tab:seed}).

\end{document}